\documentclass{article}

\PassOptionsToPackage{numbers, compress}{natbib}

\usepackage[preprint]{neurips_2021}

\usepackage[utf8]{inputenc} 
\usepackage[T1]{fontenc}    
\usepackage{hyperref}       
\usepackage{url}            
\usepackage{booktabs}       
\usepackage{amsfonts}       
\usepackage{nicefrac}       
\usepackage{microtype}      
\usepackage{xcolor}         

\AtBeginDocument{%
  \providecommand\BibTeX{{%
    \normalfont B\kern-0.5em{\scshape i\kern-0.25em b}\kern-0.8em\TeX}}}

\usepackage{amsmath}
\usepackage{amssymb}
\usepackage{subcaption}
\usepackage{comment}
\usepackage{placeins}

\usepackage{booktabs}
\usepackage{multirow}
\usepackage{adjustbox}

\usepackage[normalem]{ulem}
\usepackage{mathtools}
\usepackage{amsthm}  

\DeclarePairedDelimiterX{\norm}[1]{\lVert}{\rVert}{#1}

\newcommand{\argmax}[1]{\underset{#1}{\operatorname{arg}\,\operatorname{max}}\;}

\newcommand{\tce}{\text{CE}}
\newcommand{\tmse}{\text{MSE}}
\newcommand{\tpred}{\text{Pred}}
\newcommand{\tkd}{\text{KD}}
\newcommand{\tkl}{\text{KL}}
\newcommand{\bx}{\mathbf{x}}

\newcommand\blfootnote[1]{%
  \begingroup
  \renewcommand\thefootnote{}\footnote{#1}%
  \addtocounter{footnote}{-1}%
  \endgroup
}

\title{Going Beyond Classification Accuracy Metrics in Model Compression}

\author{
Vinu Joseph\textsuperscript{4}\footnotemark \And
Shoaib Ahmed Siddiqui\textsuperscript{2,3}\footnotemark[1] \And
Aditya Bhaskara\textsuperscript{1} \And
Ganesh Gopalakrishnan\textsuperscript{1}  \And
Saurav Muralidharan\textsuperscript{4}  \And
Michael Garland\textsuperscript{4} \And
Sheraz Ahmed\textsuperscript{3} \And
Andreas Dengel\textsuperscript{2,3} \And
\small \textsuperscript{1} \texttt{University of Utah} \qquad \small \textsuperscript{2} \texttt{TU Kaiserslautern} \\\small  \textsuperscript{3} \texttt{German Research Center for Artificial Intelligence (DFKI)} \qquad 
\small  \textsuperscript{4} \texttt{NVIDIA}
\vspace{-.5em}
}

\begin{document}

\maketitle

\blfootnote{*Joint first authors. Correspondence to Vinu Joseph: \texttt{vinuj@nvidia.com}.}

\begin{abstract}
With the rise in edge-computing devices, there has been an increasing demand to deploy energy and resource-efficient models. A large body of research has been devoted to developing methods that can reduce the size of the model considerably without affecting the standard metrics such as top-1 accuracy. However, these pruning approaches tend to result in a significant mismatch in other metrics such as fairness across classes and explainability.
To combat such misalignment, we propose a novel multi-part loss function inspired by the knowledge-distillation literature. Through extensive  experiments, we demonstrate the effectiveness of our approach across different compression algorithms, architectures, tasks as well as datasets. In particular, we obtain up to $4.1\times$ reduction in the number of prediction mismatches between the compressed and reference models, and up to $5.7\times$ in cases where the reference model makes the correct prediction; all while making no changes to the compression algorithm, and minor modifications to the loss function. Furthermore, we demonstrate how inducing simple alignment between the predictions of the models naturally improves the alignment on other metrics including fairness and attributions. Our framework can thus serve as a simple plug-and-play component for compression algorithms in the future.
\end{abstract}
\section{Introduction}
\label{sec:intro}

Modern deep learning owes much of its success to the ability to train large models by leveraging datasets of ever increasing size~\cite{waymo2019large}. The best performing models for computer vision~\cite{xie2020self} and natural language processing applications~\cite{devlin2018bert} tend to have tens to hundreds of layers and hundreds of millions of parameters. However, an increasing number of applications, including autonomous driving~\cite{teslacrash17}, surveillance~\cite{yang2018low}, and voice assistance systems~\cite{alam2020survey}, demand ML models that can be deployed on low-power and low-resource devices, and typically have strong latency requirements \cite{dennis2020edgeml}. 

In such applications, the idea of {\em model compression} has gained popularity. At a high level, model compression involves taking a {\em reference (uncompressed) model} and producing a {\em compressed model} that is lightweight in terms of computational requirements, while being functionally equivalent to the reference model (i.e., produces the same classification outputs on all inputs).
Model compression has been studied extensively in vision, speech, as well as other domains~\cite{cheng2017survey}, leveraging techniques such as 
structured weight pruning~\cite{mccarley2020structured,wang2019structured,frankle2018lottery}, quantization~\cite{zhu2016trained,gong2014compressing} and low-rank factorization~\cite{kossaifi2020factorized}.

\begin{figure}[t]
    \centering
    \includegraphics[width=\linewidth]{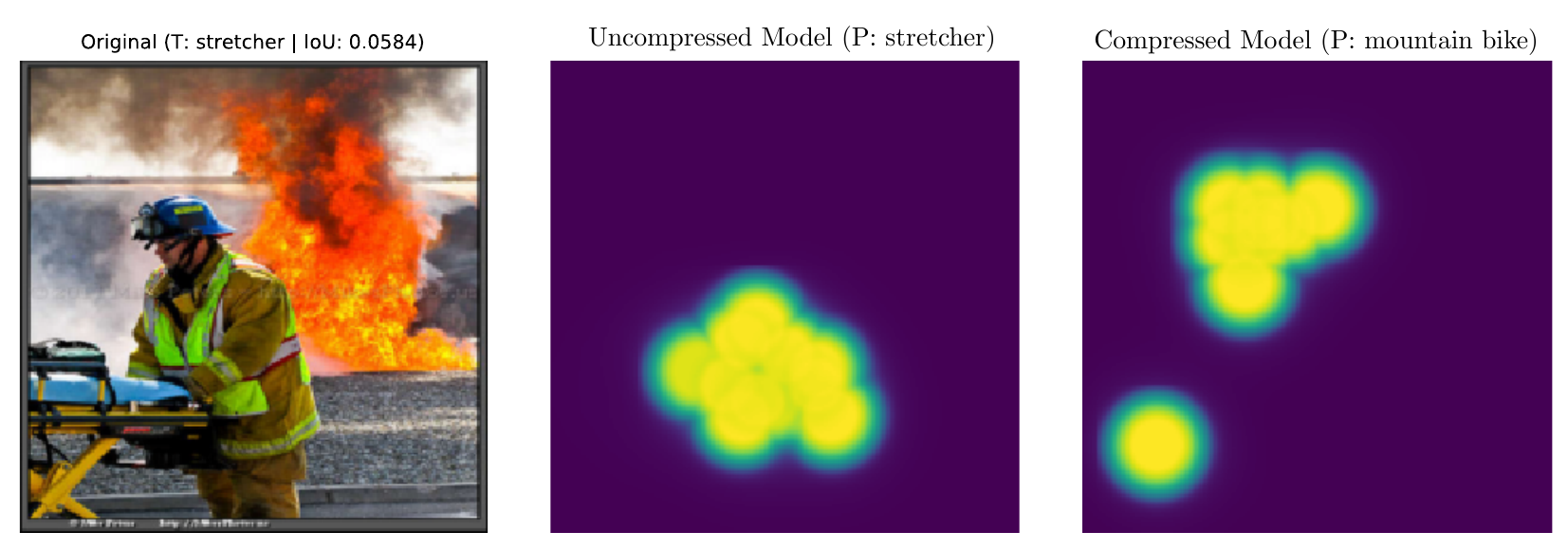}
    \caption{\small Attribution mismatch for reference and compressed models. Here, T and P refer to the target and predicted labels, respectively.}
    \label{fig:intro_new}
\end{figure}

Most known methods for compression have two steps: the first is to make structural modifications to the network (by removing connections or by creating a compressed architecture), and the second is to fine-tune the parameters to potentially regain some of the accuracy loss. The simplest and the most common way for accuracy recovery is to retrain the model using the original classification objective i.e. cross entropy loss. 
Let us denote the output of the network in the form of log-probabilities for different classes (thus it is a vector in $\mathbb{R}^C$, where $C$ is the number of classes) as $\Phi (\mathbf{x}; \mathcal{W})$ parameterized by the set of weights $\mathcal{W}$ on an input $\mathbf{x}$. If we represent the parameters of a compressed network as $\overline{\mathcal{W}}$, then the cross entropy loss takes the form:
\begin{align}
    \mathcal{L}_{\tce} = \frac{1}{N} \sum_{i=1}^{N} & \tce (\sigma(\Phi(\mathbf{x}_i; \overline{\mathcal{W}})), y_i). 
    \label{eq:loss-ce}
\end{align}
Here, the summation runs over all the training examples $\bx_i$, and $y_i$ denotes the training label for $\bx_i$. $\tce$ here represents the cross entropy loss while $\sigma$ represents the softmax function.
Cross entropy based fine-tuning is a core component of almost all state-of-the-art compression algorithms, and it often succeeds in recovering the top-$k$ accuracy even after significant compression~\cite{renda2020rewind,li2020group}.

Despite its success, however, cross entropy-based fine-tuning has its limitations. For instance, it has been shown to have a disparate class-wise impact,  i.e., the classification accuracy of some classes may be affected much more than others~\cite{8711062}. This could lead to unfair outcomes such as ethnicity-based discrimination in facial or speech recognition~\cite{buolamwini2018gender}. 
Recent work has investigated this effect in detail, leading to a push towards pruning {\em responsibly}~\cite{paganini2020prune}.

Underlying these problems is a more fundamental issue: current pruning methods can considerably alter the functional form of a classifier, as the fine-tuning objective only encourages the predictions to match the ground truth labels, without caring about the predictions of the reference (i.e., uncompressed) model. 
This misalignment with the reference also manifests in other aspects such as interpretability where the compressed model focuses on entirely different features as compared to the uncompressed model.
One such example is visualized in Figure~\ref{fig:intro_new}, where the compressed and the uncompressed models entirely disagree in terms of the features they use to reach their predictions.

The motivation behind our work can be summarized as follows. Modern neural networks are optimized for a variety of semantic metrics, such as fairness, interpretability, privacy-preserving feature selection, etc. But deploying these networks in edge and other applications requires a compression step, which can considerably affect these semantic metrics. Instead of re-doing the entire optimization in a compression pipeline,  {\em can we develop compression schemes that automatically ensure alignment between the compressed and the uncompressed models at a more semantic level?}
This question is challenging because networks of different sizes are difficult to compare. Moreover, notions such as fairness and feature similarity can be formalized in many distinct ways (for example, see~\cite{du2020fairness}). The problem also inherently involves the simultaneous optimization of multiple objectives, which is challenging in itself. 

\subsection{Our Contributions}
Our goals in the paper are two-fold: (a) to develop a general approach to minimize the misalignment between the compressed and the uncompressed models, and (b) to identify metrics that can yield insight into how different compression schemes perform in terms of preserving class-wise accuracy and {\em feature similarity} between the models. 

\vspace{-6pt}
\paragraph{Alignment via knowledge distillation.} Our high level approach towards goal (a) is knowledge distillation~\cite{hinton2015distilling}. 
Distilling knowledge from an uncompressed model to a compressed model is one of the most effective ways of obtaining a high accuracy compressed model. 
The distillation objective encourages the classification probabilities (for all classes) of the compressed and the uncompressed model to align.  Consider the temperature augmented softmax function, $\sigma(\mathbf{z}, T) = \frac {e^{\mathbf{z} / T}} {\sum_{j=1}^{C} e^{\mathbf{z}_i / T}}$, where $T$ controls the extent of {\em diffusion} of the distribution.  
%
Then the objective that knowledge distillation minimizes is:
\begin{equation}
    \mathcal{L}_{\tkd} = \frac{1}{N} \sum_{i=1}^{N} \tkl \left( \sigma (\Phi (\mathbf{x}_i; \overline{\mathcal{W}})), \sigma (\Phi (\mathbf{x}_i; \mathcal{W}), T) \right),
\end{equation}
where $\tkl$ represents the KL-divergence between the two distributions. By tuning the temperature parameter, the loss function can be used to control how much significance to assign to the ``top'' class (one with the maximum probability) relative to the other classes. 

Distillation is part of a more general ``teacher-student'' paradigm for training networks. Variants of distillation, such as {\em logit pairing} have also been shown to be practically effective; for instance, Kannan et al.~\cite{kannan2018adversarial} show that the following objective has the added advantage of not relying on tuning a temperature parameter, while achieving similar performance.
\begin{equation}
    \mathcal{L}_{\tmse} = \frac{1}{N} \sum_{i=1}^{N} \norm{ \Phi (\mathbf{x}_i; \overline{\mathcal{W}}) - \Phi (\mathbf{x}_i; \mathcal{W})}^2.
    \label{eq:logit_pairing_loss}
\end{equation}
Another variant is one in which we use cross entropy as in~\eqref{eq:loss-ce}, but match the classification with the output of the uncompressed model (hard-distillation), i.e.
\begin{align}
    \mathcal{L}_{\tce_{\tpred}} = \frac{1}{N} \sum_{i=1}^{N}  \tce (\sigma(\Phi(\mathbf{x}_i; \overline{\mathcal{W}})), y_i'),  ~~
    \text{ where } y_i' := \argmax{j} \Phi(\mathbf{x}_i; \mathcal{W})_{j}. \label{eq:loss-ce-pred}
\end{align}
All of the loss functions above promote alignment between the reference and the compressed models. Thus the main technical question is, {\em how do we combine different loss objectives} into a compression pipeline?
To address this, we present three different algorithms with varying levels of tuning, and study their performance on various semantic metrics in Section~\ref{sec:method}.

\vspace{-6pt}
\paragraph{Semantic metrics for model compression.}  As discussed above, it is important in applications to go beyond the top-$k$ accuracy when studying compression schemes. We study metrics that can capture the ``functional equivalence'' of models, as well as feature similarity. 
A quantitative measure of functional equivalence (i.e., the reference and compressed models mapping an input to the same output) is to measure the number of test examples in which the classifications differ. Such inputs have been termed {\em compression impacted exemplars} (CIEs) in recent work by Hooker et al. (2020)~\cite{hooker2020characterising}.\footnote{\label{footnote:cie-cie}Note that~\cite{hooker2020characterising} used the slightly different name of Compression {\em Identified} Exemplars; our modification is done to ensure compatibility with other notions that we define.} Note that the number of CIEs can be large {\em in spite of the reference and compressed models having the same top-$k$ accuracy} as the models may make mistakes on different inputs. In settings where the reference model has been optimized for a different objective, e.g., ensuring fairness across classes, minimizing the number of CIEs enables us to obtain similar guarantees for the compressed model. We introduce two slight variants of CIEs (denoted CIE-U and CIP) and use them to compare compression schemes (see Section~\ref{sec:metrics_misalignment}).
As a second metric, we measure the class-level impact of compression schemes. As discussed earlier, compression can lead to a loss in classification accuracy for some classes more than others and the metrics we will define help us quantify the disparate impact. 
The third metric we propose aims to capture the similarity between the {\em features} of the original and the compressed model. Our goal is to capture the notion that the models ``perceive'' the same features from images. We do this by considering saliency maps (or attributions, studied extensively as a tool for model interpretability~\cite{kindermans2019reliability}), and defining an IoU (intersection over union) measure over the saliency maps. The details of all the three metrics are discussed in  Section~\ref{sec:metrics_misalignment}.

\vspace{-6pt}
\paragraph{Experimental evaluation.}  The setup we have discussed so far involves multiple variables, including different compression algorithms, network architectures, distillation objectives (and various ways to combine them), and various evaluation metrics for capturing semantic similarities between the reference and compressed networks. In Section~\ref{sec:results}, we present the results of an extensive set of experiments that cover all the important combinations. 
We arrive at the interesting conclusion that optimizing the logit-pairing objective from Eq.~\eqref{eq:logit_pairing_loss} seems to have the most impact across all the metrics we study. We also observe correlations among metrics: techniques that help reduce CIEs improve fairness metrics as well, and improved IoU in attributions also correlates with reduced CIEs. 


\section{Model Compression Baselines}
\label{sec:related}



%


Model compression is now a rich area of research with several algorithmic techniques and fine-tuning procedures~\cite{cheng2017survey,cheng2018recent}. As baselines for our work, we have used compression methods from~\cite{joseph2019condensa, li2020group} that allow us to easily incorporate novel distillation objectives. 

\textit{Group Sparsity} is a recent state-of-the-art technique for model compression. In its finetuning step, it incorporates a knowledge distillation term with $\tce$~\cite{li2020group}. 
This serves as a good baseline for our experiments.
The general form of the optimization problem used in group sparsity is:
\begin{equation}
\min _{\mathcal{W}} \mathcal{L}(y, \Phi(\mathbf{x} ; \mathcal{W}))+\mu \mathcal{D}(\mathcal{W})+\lambda \mathcal{R}(\mathcal{W}),
\label{eq:main:loss1}
\end{equation}
\noindent where $\mathcal{D}$ refers to weight decay and $\mathcal{R}$ refers to an appropriately defined sparsity regularization term. 
In this paper, we build upon this approach and adapt the loss function shown in Equation~\ref{eq:main:loss1} to better encourage alignment between the compressed and reference models.


We also evaluate our proposed approach on the recently-proposed \textit{REWIND} pruning algorithm~\cite{renda2020rewind}, which is relatively simple but works surprisingly well in practice.
In this paper, we consider the simplest variant of rewind where the whole model is retrained after every pruning step using the same schedule that was used to train the initial model. As recommended by Renda et al.~\cite{renda2020rewind}, the sparsity enforced at each pruning step increases by a fixed factor (20\% in our experiments).

\section{Methodology}
\label{sec:method}

We now delve into the details of our methods. We start by defining metrics that capture misalignment between models and then propose techniques to improve the alignment.

\subsection{Metrics for Model Misalignment}
\label{sec:metrics_misalignment}

Misalignment between models can occur at several different semantic levels ranging from basic misalignment at the prediction-level to misalignment in metrics such as model fairness or interpretability.
We will now briefly discuss how these different forms of misalignment can be quantified via useful metrics that can be reported as part of compression methods in the future.

\paragraph{Compression Impacted Exemplars (CIEs).}
The simplest form of misalignment that can be captured among models is in terms of their predictions. Hooker et al.~\cite{hooker2020characterising} introduced the notion of compression impacted exemplars (CIE): test examples that the compressed and uncompressed models classify {\em differently}. Having near-zero number of CIEs
demonstrates the functional equivalence of models. In our experiments, we measure the number of CIEs, and also introduce two variants. Note that CIEs are more important when the reference model makes the {\em correct} (agreeing with ground truth) prediction, because this means that the compressed models gets that example wrong. We denote such inputs by {\em CIE-U}, and report the number of such inputs as well. 
%
%
%
CIEs are useful when classifying entire images, but the general notion can be applied in other settings as well. In the context of semantic segmentation ---a classification task that assigns a class to every pixel in an image--- we extend the notion of CIEs to mean {\em pixels} for which the reference and compressed models disagree in classification. We term these {\em Compression Impacted Pixels}, denoted CIPs. Using a semantic segmentation task on brain MRI images, we demonstrate (in Section~\ref{sec:results}) the efficacy of our techniques in reducing the number of CIPs. 

\newcommand{\eps}{\varepsilon}
\paragraph{Model Fairness.}
Ensuring that a machine learning model is {\em fair} is a challenging problem that has seen a variety of approaches (see, e.g., the survey of~\cite{du2020fairness}). 
The notion of fairness is often application-specific, making it hard to quantify. In the context of multi-class classification of images, ``class level'' fairness metrics have been used~\cite{paganini2020prune}. The first metric we report is the $max-min$ accuracy, defined as follows: suppose $\eps_i$ denotes the fraction of inputs of class $i$ that are misclassified. Then $max-min$ is defined as $\max_{i \in \mathcal{C}} \eps_i - \min_{i \in \mathcal{C}} \eps_i$, where $\mathcal{C}$ is the set of classes.
While the metric captures the notion of one class incurring more error than others, it can depend too much on outlier classes. To better understand class-level impact of compression, we also compute the distribution of differences between the uncompressed and the compressed accuracy. In this case, negative values indicate an increase in the accuracy of the compressed model for a particular class while positive values highlight a reduction.

\paragraph{Model Interpretability.}
Explainability of deep neural networks is a rich area of research ~\cite{barocas-hardt-narayanan, xie2020explainable}. Although various ways of quantifying explainability have been proposed, one of the simplest and the most accessible one is of attributions~\cite{kindermans2019reliability} (also known as saliency maps). Attributions assign a measure of importance to each pixel or attribute of the input that indicates how much the final prediction relies on the values of these attributes.

We hypothesize that similarity in attributions implies {\em feature similarity} between the reference and compressed models (they use similar features for classification). We thus propose a soft IoU (intersection over union) metric defined using attribution maps obtained from the compressed and uncompressed models (and average this over input images): $iou(\mathbf{x}, \mathbf{y}) = min(\mathbf{x}, \mathbf{y}) / max(\mathbf{x}, \mathbf{y})$ where both $min$ and $max$ computes the element-wise minimum and maximum.  We note that the idea of matching attributions in compression was independently pursued in the recent work of~\cite{park2020attribution}. 

\subsection{Minimizing Model Misalignment}
\label{sec:minimize_misalignment}

Given a reference model $\mathcal{W}$, we wish to obtain a compressed model $\overline{\mathcal{W}}$ that: (i) obtains the almost the same accuracy as $\mathcal{W}$, and (ii) aligns with $\mathcal{W}$ in all the metrics discussed above. 

As described in Section~\ref{sec:intro}, knowledge distillation lies at the core of our approach. The different distillation objectives defined in Eq.~\eqref{eq:loss-ce},~\eqref{eq:logit_pairing_loss},~\eqref{eq:loss-ce-pred} all provide loss terms whose minimization can improve alignment between the models. Thus the challenge is to deal with this multi-objective optimization.

The issue of optimizing multiple objectives arises in many contexts in learning, and several works have tackled the question of how best to combine loss functions~\cite{huang2019addressing,barron2019general,chen2018gradnorm}. The natural approach is to consider a linear combination of the losses:
\begin{equation}
    \mathcal{L} = \alpha \cdot \mathcal{L}_{\tce} + \beta \cdot \mathcal{L}_{\tmse}  + \gamma \cdot \mathcal{L}_{\tce_{\tpred}}.
    \label{eq:loss}
\end{equation}
The question then is to find the right methods to tune the hyper-parameters $\alpha$, $\beta$ and $\gamma$. While a lot of the prior work combines loss functions either using ad-hoc or equal weights, there have been some systematic approaches to adjust
the weights in an automated manner. 
These methods often require defining new loss functions~\cite{barron2019general} or changing the optimization procedure~\cite{chen2018gradnorm}. 

In our work, we experiment with three different ways of setting the hyper-parameters $\alpha$, $\beta$, and $\gamma$ in Eq.~\ref{eq:loss}: \textsc{Uniform}, \textsc{Learnable}, and \textsc{SoftAdapt}. To better understand the contributions of the individual loss terms to the final model, we also evaluate the 7 possible subsets  obtained using the three loss terms in Eq.~\ref{eq:loss}. These subsets correspond to setting some of the loss weights to zero, while optimizing the rest of the weights using the three methods we describe.

\noindent \textsc{Uniform}. Here, we assign a uniform weight to each of the selected loss terms. The weight is equally divided among the number of loss terms present. i.e., for subsets containing 1, 2, and 3 loss terms, the weights are $1.0$, $0.5$ and $0.33$ respectively.

\noindent \textsc{Learnable}. In the learnable variant, we treat the weights $\alpha, \beta, \gamma$ as parameters of the model, and then optimize them as standard parameters using gradient descent. Since the gradient points in the steepest direction, using gradient descent can become biased towards different terms depending on their relative magnitude. To prevent weights from becoming zero, we introduce a weight decay term and project the weights to sum to 1. 

\noindent \textsc{SoftAdapt}.
Proposed by Heyderi et al.~\cite{heydari2019softadapt}, \textsc{SoftAdapt} is a method to automatically tune the weights of a multi-part loss function. It can be tuned to assign the maximum weight to either the best-performing or the worst-performing loss based on the change in the loss functions and a parameter $\eta$.

\section{Experiments}
\label{sec:results}

\begin{table*}[!t]
\centering
\begin{adjustbox}{width=1.0\linewidth,center}
\begin{tabular}{lllrrrr}
\toprule
   Dataset & Network (Sparsity) &                Loss function &  Uncompressed Acc. &  Compressed Acc. &   \#CIEs &  \#CIE-Us \\
\midrule
  &    &                          CE-Dist. (Baseline, \cite{li2020group}) &                $76.78\%$ &              $74.95\%$ &   $1925$ &     $717$ \\
  &   RESNET-164$(0.56)$ &                UNIFORM (MSE) &                $76.78\%$ &              $76.58\%$ &   $\mathbf{1092(1.8\times)}$ &     $\mathbf{335(2.1\times)}$ \\
  &    &                          UNIFORM (CE, MSE) &                $76.78\%$ &              $76.62\%$ &   $1237(1.6\times)$ &     $386(1.9\times)$ \\
  \cline{2-7}
  &     &                         CE-Dist. (Baseline, \cite{li2020group}) &               $68.83\%$ &              $66.96\%$ &   $1903$ &          $639$ \\
  &    RESNET-$20(0.33)$ &                UNIFORM (MSE) &                $68.83\%$ &              $68.87\%$ &    $\mathbf{465(4.1\times)}$ &     $\mathbf{113(5.7\times)}$ \\
 CIFAR-100 &     &                UNIFORM (CE, MSE) &                $68.83\%$ &              $69.13\%$ &    $717(2.7\times)$ &     $184(3.5\times)$ \\
 \cline{2-7}
  &   &                           CE-Dist. (Baseline, \cite{li2020group}) &                $76.87\%$ &              $74.40\%$ &   $2092$ &     $788$ \\
  &  RESNEXT-$164(0.48)$ &                UNIFORM (MSE) &                $76.87\%$ &              $76.15\%$ &   $\mathbf{1327(1.6\times)}$ &     $\mathbf{439(1.8\times)}$ \\
  &   &                           UNIFORM (CE, MSE) &                $76.87\%$ &              $75.95\%$ &   $1523(1.4\times)$ &     $526(1.5\times)$ \\
  \cline{2-7}
  &    &                           CE-Dist. (Baseline, \cite{li2020group}) &                $71.95\%$ &              $70.98\%$ &   $1609$ &    $535$ \\
  &   RESNEXT-$20(0.54)$ &                UNIFORM (MSE) &                $71.95\%$ &              $72.13\%$ &    $\mathbf{601(2.7\times)}$ &     $\mathbf{161(3.3\times)}$ \\
  &    &            UNIFORM (CE, MSE) &                $71.95\%$ &              $72.16\%$ &    $792(2\times)$ &     $202(2.6\times)$ \\
\midrule
   &      &                          CE-Dist. (Baseline, \cite{li2020group}) &                $94.62\%$ &              $92.90\%$ &    $550$ &     $317$ \\
   &     DENSENET$(0.56)$ &                UNIFORM (MSE) &                $94.62\%$ &              $93.37\%$ &    $496(1.1\times)$ &     $274(1.2\times)$ \\
   &      &            UNIFORM (CE, MSE) &                $94.62\%$ &              $94.40\%$ &    $\mathbf{301(1.8\times)}$ &     $\mathbf{135(2.3\times)}$ \\
   \cline{2-7}
   &    &                           CE-Dist. (Baseline, \cite{li2020group}) &                $95.03\%$ &              $93.71\%$ &    $466$ &     $266$ \\
   &   RESNET-$164(0.54)$ &                UNIFORM (MSE) &                $95.03\%$ &              $94.19\%$ &    $381(1.2\times)$ &     $208(1.3\times)$ \\
   &    &            UNIFORM (CE, MSE) &                $95.03\%$ &              $94.33\%$ &    $\mathbf{354(1.3\times)}$ &     $\mathbf{189(1.4\times)}$ \\
   \cline{2-7}
   &     &                           CE-Dist. (Baseline, \cite{li2020group}) &                $92.54\%$ &              $90.54\%$ &    $657$ &     $381$ \\
   &    RESNET-$20(0.45)$ &                UNIFORM (MSE) &                $92.54\%$ &              $92.28\%$ &    $396(1.7\times)$ &     $176(2.2\times)$ \\
  CIFAR-10 &     &            UNIFORM (CE, MSE) &                $92.54\%$ &              $92.46\%$ &    $\mathbf{223(2.9\times)}$ &     $\mathbf{101(3.8\times)}$ \\
  \cline{2-7}
   &     &                           CE-Dist. (Baseline, \cite{li2020group}) &                $93.09\%$ &              $91.73\%$ &    $589$ &     $323$ \\
   &    RESNET-$56(0.5)$ &                  UNIFORM (MSE) &                    $93.09\%$ &              $91.73\%$ &    $572(1.0\times)$ &     $311(1.0\times)$ \\
   &     &                           UNIFORM (CE, MSE) &                $93.09\%$ &              $92.08\%$ &    $530(1.1\times)$ &     $\mathbf{278(1.2\times)}$ \\
   \cline{2-7}
   &   &                           CE-Dist. (Baseline, \cite{li2020group}) &                $95.18\%$ &              $93.74\%$ &    $472$ &     $276$ \\
   &  RESNEXT-$164(0.44)$ &                UNIFORM (MSE) &                    $95.18\%$ &              $92.87\%$ &    $576(0.8\times)$ &     $373(0.7\times)$ \\
   &   &                           UNIFORM (CE, MSE) &                $95.18\%$ &              $93.83\%$ &    $462(1.0\times)$ &     $269(1.0\times)$ \\
   \cline{2-7}
   &    &                           CE-Dist. (Baseline, \cite{li2020group}) &                $92.54\%$ &              $90.92\%$ &    $783$ &     $422$ \\
   &   RESNEXT-$20(0.60)$ &                UNIFORM (MSE) &                $92.54\%$ &              $91.37\%$ &    $674(1.2\times)$ &     $345(1.2\times)$ \\
   &    &            UNIFORM (CE, MSE) &                $92.54\%$ &              $92.42\%$ &    $\mathbf{464(1.7\times)}$ &     $\mathbf{197(2.1\times)}$ \\
\midrule
   &     &                           CE-Dist. (Baseline, \cite{li2020group}) &                $76.01\%$ &              $76.35\%$ &   $4491$ &    $1185$ \\
  IMAGENET &    RESNET-$50(0.47)$ &                UNIFORM (MSE) &                $76.01\%$ &              $76.13\%$ &   $1890(\mathbf{2.4\times})$ &     $\mathbf{500(2.4\times)}$ \\
   &     &            UNIFORM (CE, MSE) &                $76.01\%$ &              $76.48\%$ &   $2911(1.5\times)$ &     $704(1.7\times)$ \\
  \midrule
  \midrule
        &          &                 UNIFORM (CE) &                 $0.8454$ &               $0.8224$ &  $38373$ &   $15856$ \\
       MRI &         UNET(0.81) &                UNIFORM (MSE) &                 $0.8454$ &               $0.8576$ &  $\mathbf{22988(1.7\times)}$ &   $13468(1.2\times)$ \\
        &          &            UNIFORM (CE, MSE) &                 0.8454 &               $0.8581$ &  $\mathbf{23183(1.7\times)}$ &   $13556(1.2\times)$ \\
\bottomrule
\end{tabular}
\end{adjustbox}
\caption{Summary of our main results. We compare the performance of our two best-performing losses and one baseline on 12 networks across the CIFAR-10/100, ImageNet and Brain MRI datasets. The entries in bold correspond to losses that perform best for that architecture+dataset combination. The numbers in parentheses in the second column denote the sparsities of the corresponding compressed networks.}
\label{tab:results}
\end{table*}

\begin{figure*}[h]
    \centering
    \includegraphics[width=0.8\linewidth]{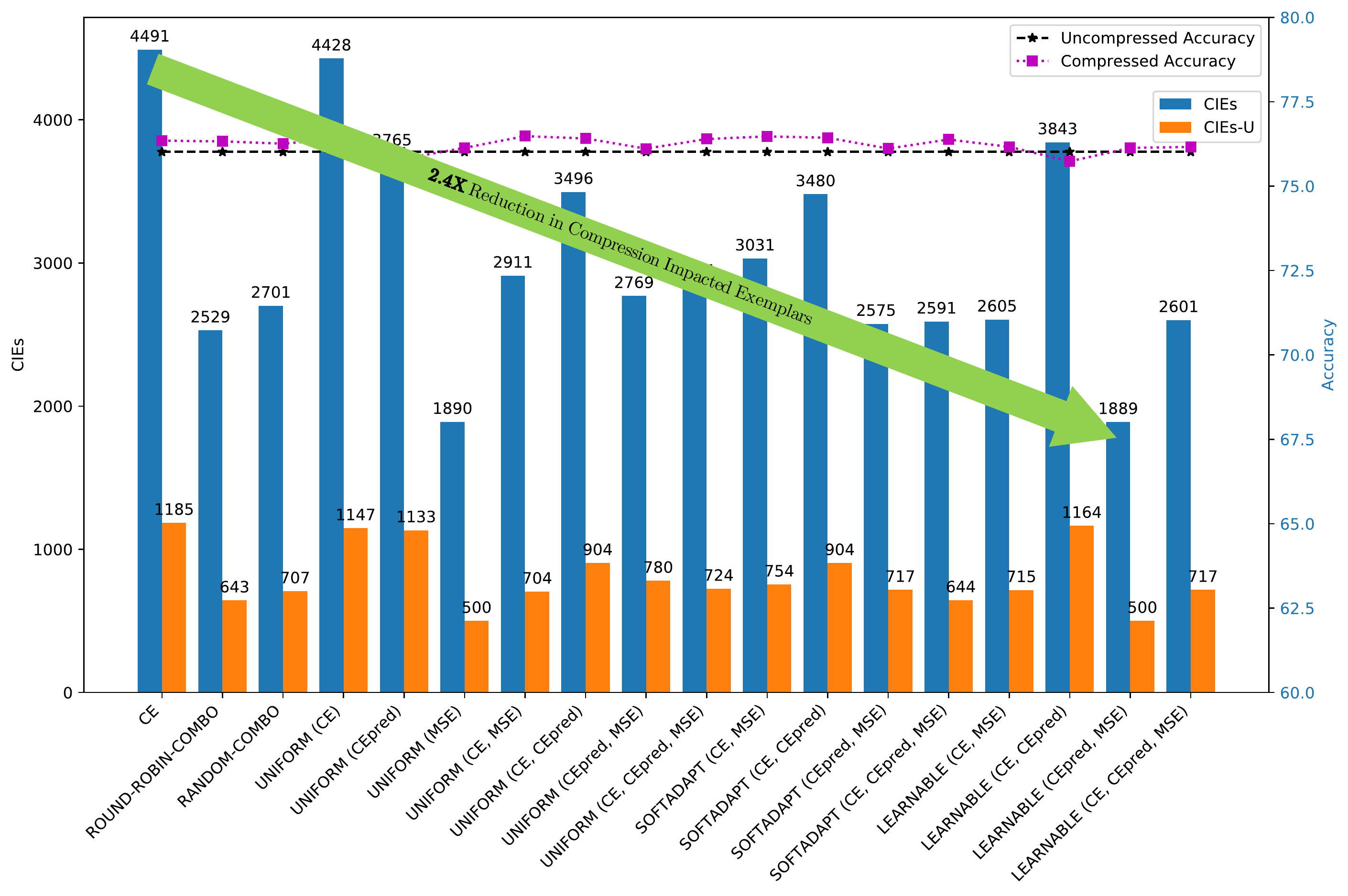}
    \caption{\small Performance of each loss+optimizer combination on ResNet50 (ImageNet)}
    \label{fig:resnet50_imagenet}
\end{figure*}

We evaluate the efficacy of the proposed label-preservation-aware loss functions on a wide range of real-world tasks, network architectures, and datasets. Specifically, we report results for image classification on the CIFAR-10/100~\cite{krizhevsky2014cifar} and ImageNet~\cite{russakovsky2015imagenet} datasets, and semantic segmentation on the Brain MRI Segmentation dataset~\cite{buda2019brainmri}. To demonstrate that our approach is not restricted to particular architectures, we experiment with a range of different network architectures, including ResNet-20 , ResNet-50 \cite{he2016deep}, ResNeXt \cite{xie2017aggregated}, ResNet-56, DenseNet 12-40 \cite{huang2017densely} and U-Net \cite{ronneberger2015u}.

As described in Section~\ref{sec:method}, we consider the CE, MSE, and $\text{CE}_{\text{PRED}}$ losses and use three algorithms to obtain optimal weights for each of these losses: \textsc{Uniform}, \textsc{Learnable}, and \textsc{SoftAdapt}. We evaluate each of these combinations and also include three additional baseline comparisons: (1) CE-Distillation, or group sparsity~\cite{li2020group} (described in detail in Section~\ref{sec:related}), (2) \textsc{Round-Robin-Combo}, which picks a loss to minimize in round-robin fashion, and (3) \textsc{Random-Combo}, which simply picks one loss at random for optimization.

\paragraph{Hyper-Parameter Settings}
For ResNet20 and ResNet56 on the CIFAR datasets, the residual
block is the basic ResBlock with two $3 \times 3$ convolutional
layers. 
For ResNet164 on CIFAR and ResNet50 on ImageNet, the residual block is a bottleneck block. ResNeXt20 and ResNeXt164 have cardinality $32$, and bottleneck width $1$.
For CIFAR, we train the reference network for $300$ epochs with SGD using
a momentum of $0.9$, weight
decay of $10^{-4}$, and batch size of $64$;
the learning rate starts with $0.1$ and decays by $10$ at epochs $150$ and $225$.
The ResNet50 model for ImageNet is obtained from the PyTorch pretrained model repository ~\cite{NEURIPS2019_9015}.
We use NVIDIA V100 GPUs to train all models.
For compression, we follow the settings described in Li et al.~\cite{li2020group}; in particular, we use $\ell_1$ regularization with a regularization factor of $2e^{-4}$, and use different learning rates for $\bf{W}$ and $\bf{A}$. The ratio between $\eta_s$ and $\eta$ is $0.01$.

\subsection{Results on CIE and CIP reduction}
Table~\ref{tab:results} summarizes our main results. Here, we compare our best-performing loss+optimizer combination with the CE-Distillation group sparsity baseline (described above) for each task, network architecture, and dataset. In the table, the first three columns denote the dataset, network (along with sparsity ratio of the compressed models) and loss function being evaluated, respectively. As shown in the Table, we demonstrate significant reductions in the number of CIEs (up to $4.1\times$) and CIE-Us (up to $5.7\times$) using label-preservation-aware loss functions while largely retaining reference top-1 accuracy. {\em Further, we notice that one of the simplest loss+optimizer combinations, namely CE and MSE with uniform weights works best in practice.} We now dive deeper into how our proposed loss functions perform for each individual task and dataset.

\subsubsection{Image Classification on CIFAR-10/100}
On CIFAR-10 and CIFAR-100, we achieve CIE reductions of up to $2.94\times$ and $4.09\times$, respectively, with a negligible drop in accuracy (less than 0.1\%) as shown in Table~\ref{tab:results}. Compared to ImageNet, uniform weights with CE and MSE losses performs best on CIFAR.

\subsection{Scaling to ImageNet Classification}

To identify the individual contribution of each of the loss terms, we performed detailed experiments over all our loss combinations on the networks and datasets shown in Table~\ref{tab:results}. Due to space restrictions, we only show results for ResNet50 (ImageNet) in this paper in Figure~\ref{fig:resnet50_imagenet}.
As shown in the figure, both \textsc{Uniform}-(MSE) and \textsc{Learnable}-(CEpred, MSE) achieve CIE and CIE-U reductions of $2.4\times$ while improving upon baseline top-1 accuracy.

\subsection{Error Bars: Measuring Variance in CIEs}


To better understand how CIEs vary with different loss term weight initialization, we evaluated 10 random runs of our loss+optimizer combinations on ResNet20 (CIFAR-100). In the supplement, we provide 
the results of this experiment in the form of a box-plot. We notice that the number of CIEs across the 10 random runs are fairly consistent for the majority of losses. We observe similar trends on other networks and datasets. 

\subsection{Application to Sensitive Domains: Semantic Segmentation on Brain MRI}

We highlight in Section~\ref{sec:method} that the notion of CIEs can naturally extended to pixels in a semantic segmentation task.
Therefore, we evaluate our approach on FLAIR MRI segmentation.
The dataset comprises of brain MRI images along with manual FLAIR abnormality segmentation masks~\cite{buda2019brainmri}.
We use a generic U-Net architecture with two output channels, and train the complete model using conventional cross-entropy loss.
We prune the model using unstructured pruning at a sparsity of 81\% and the rewind algorithm~\cite{renda2020rewind}.

\begin{figure*}[t]
    \centering
    \begin{subfigure}[b]{0.3\textwidth}
    \includegraphics[width=\textwidth]{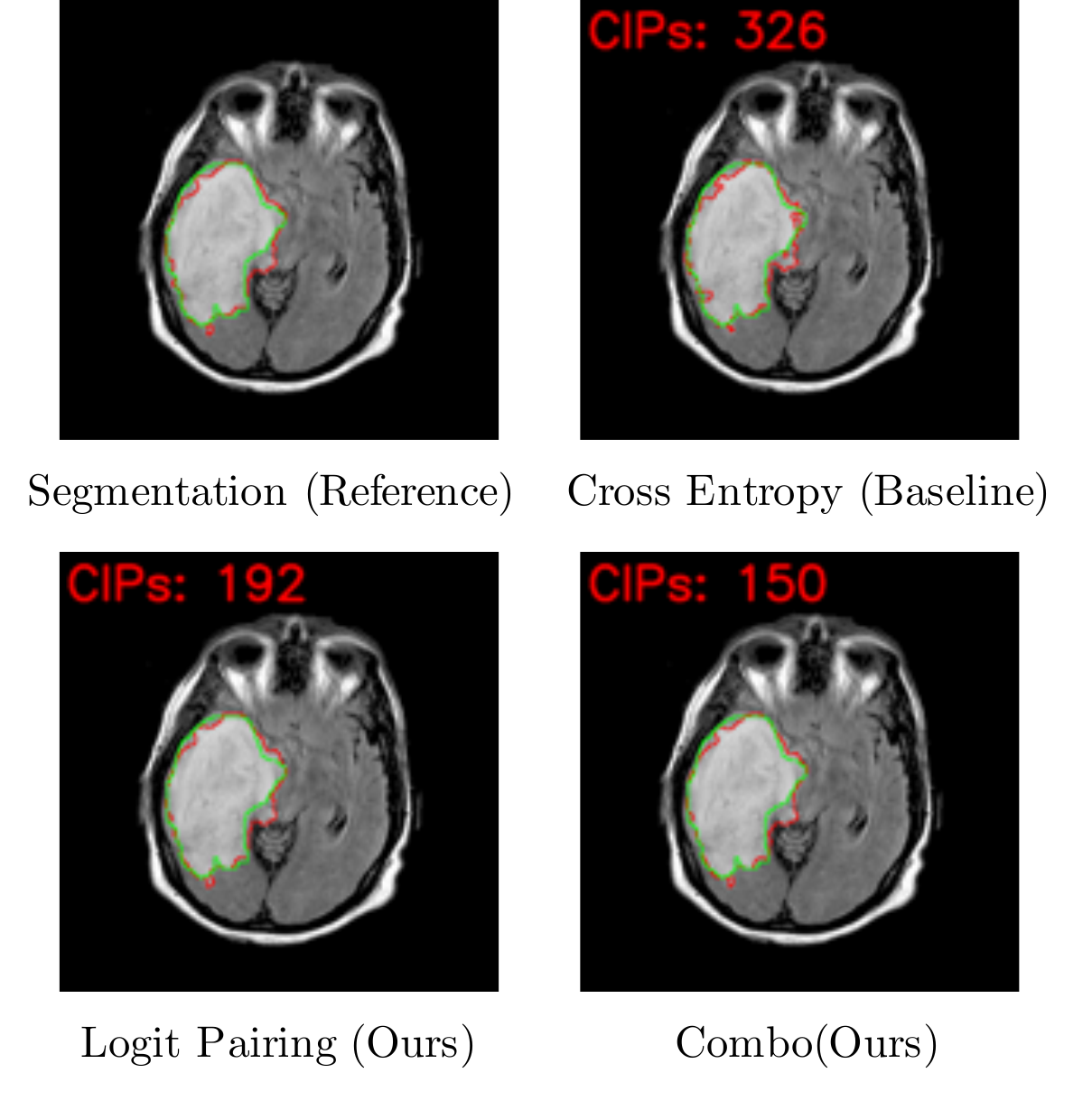}
    \end{subfigure}
    \begin{subfigure}[b]{0.4\textwidth}
    \includegraphics[width=\textwidth]{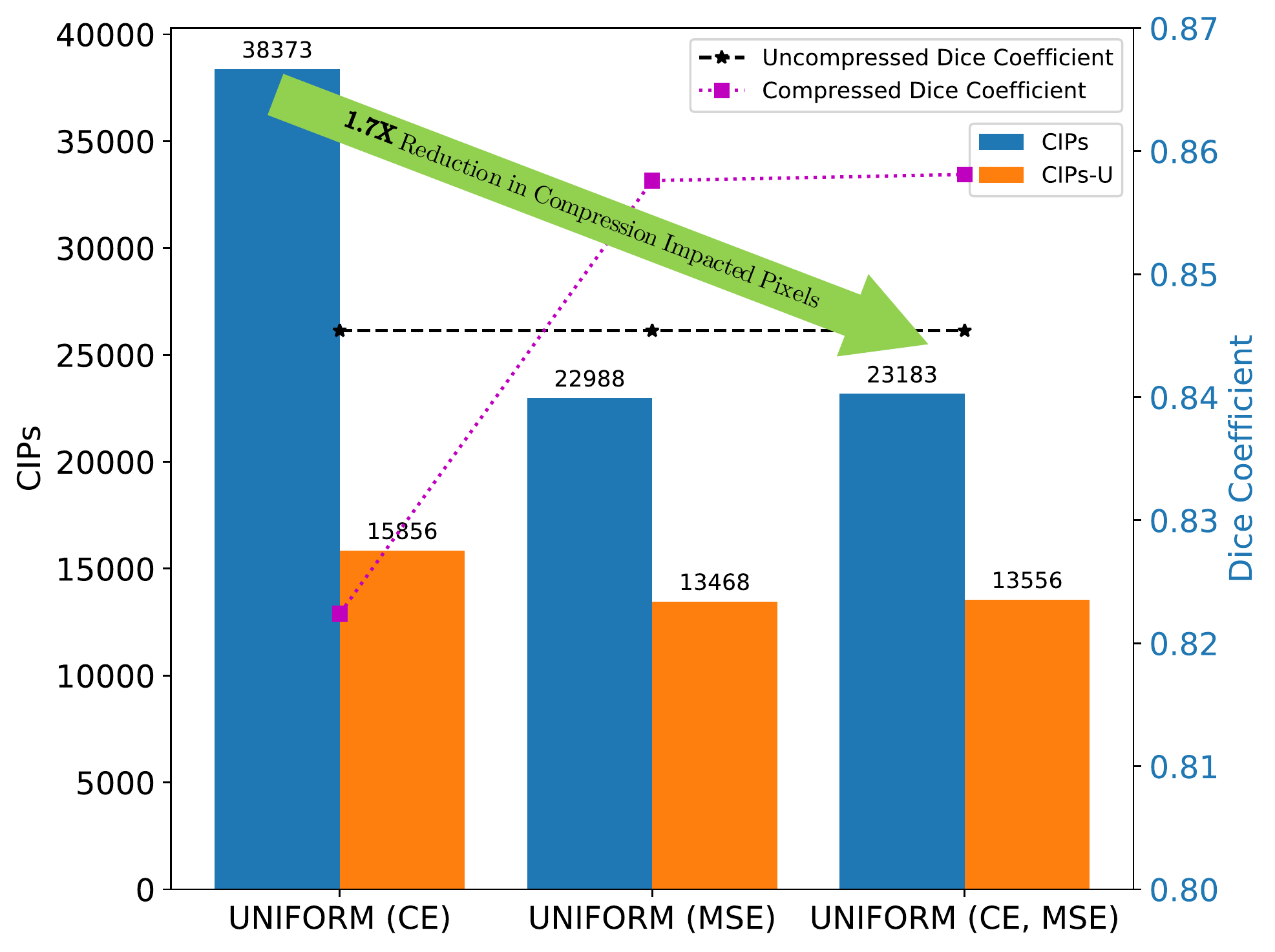}
    \end{subfigure}
    \caption{Left: CIPs for Brain MRI FLAIR Segmentation. CIPs are pixels classified differently in the original and compressed models. Red contour represents the prediction while the green contour represents the ground-truth. Right: comparison of CIPs for different loss functions.}
    \label{fig:cips_results}
\end{figure*}

The results are visualized in Fig.~\ref{fig:cips_results}.
Similar to classification, we notice a drastic reduction in the number of CIPs when using logit pairing (MSE). We also see a positive impact of including CE along with MSE on the dice coefficient, as the model only focuses on reducing CIP-Us. This indicates that using the proposed label-preservation-aware loss functions can naturally mitigate the impact of compression on the functional form of the classifier even for tasks beyond image classification.
We also include some specific qualitative examples of CIP reduction on brain MRI images in Figure~\ref{fig:cips_results}.

\subsection{Fairness Preservation}
\label{sec:fairness}

The class-level accuracy for ResNet-20 (CIFAR-10) along with per-class CIEs is shown in Figure~\ref{fig:intro} (bottom). We notice from the figure that some classes are more severely impacted than others (for example, class ID-3 (cats) in CIFAR-10).
This indicates that there is desperate impact of compression on some classes in particular.
This is against the notion of fairness.

\begin{figure*}[t]
    \centering
    \begin{subfigure}[b]{0.45\textwidth}
        \includegraphics[width=\textwidth]{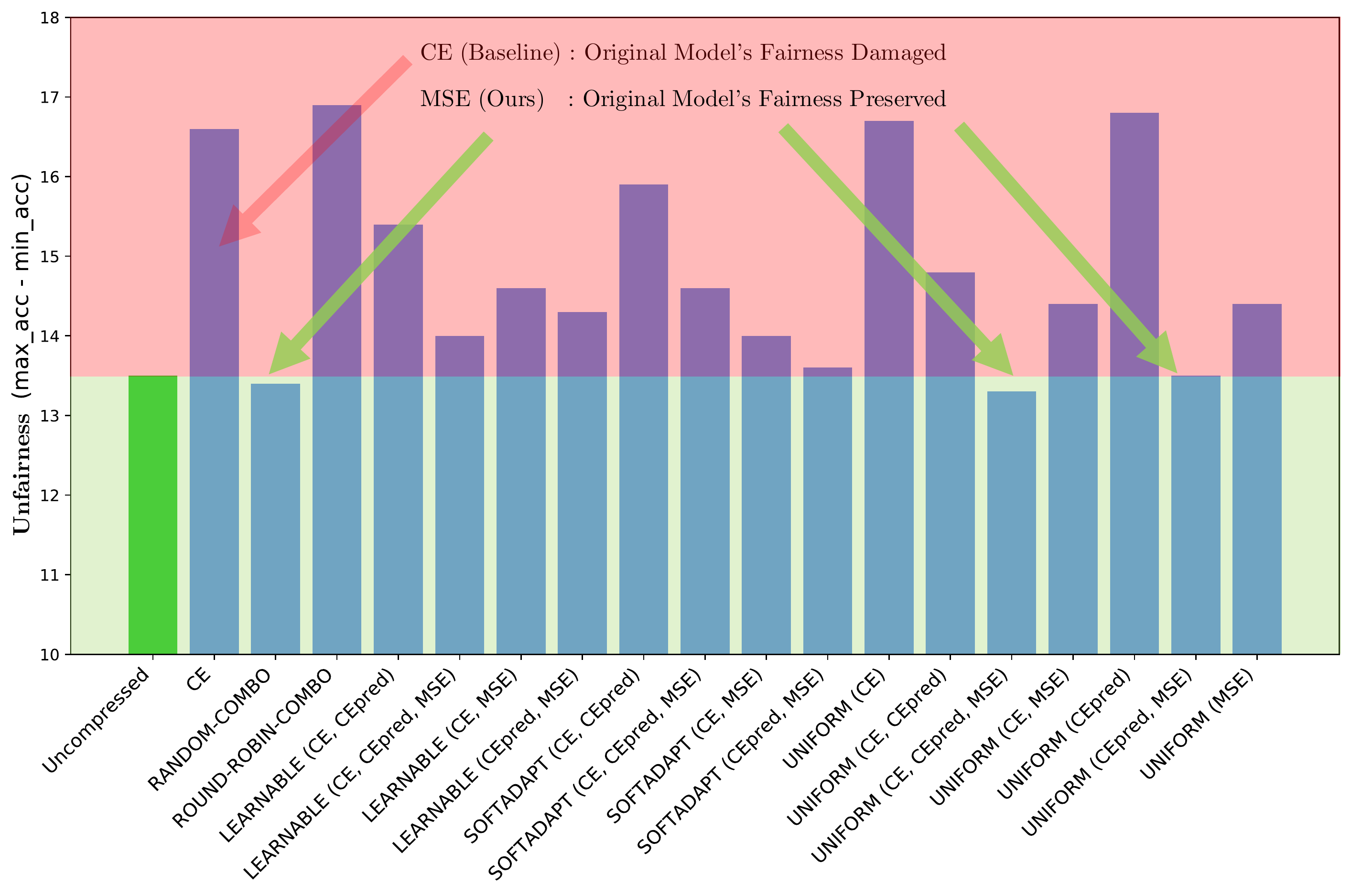}
        \label{fig:fairness_results_paganini}
    \end{subfigure}
    \begin{subfigure}[b]{0.36\textwidth}
    \includegraphics[width=\textwidth]{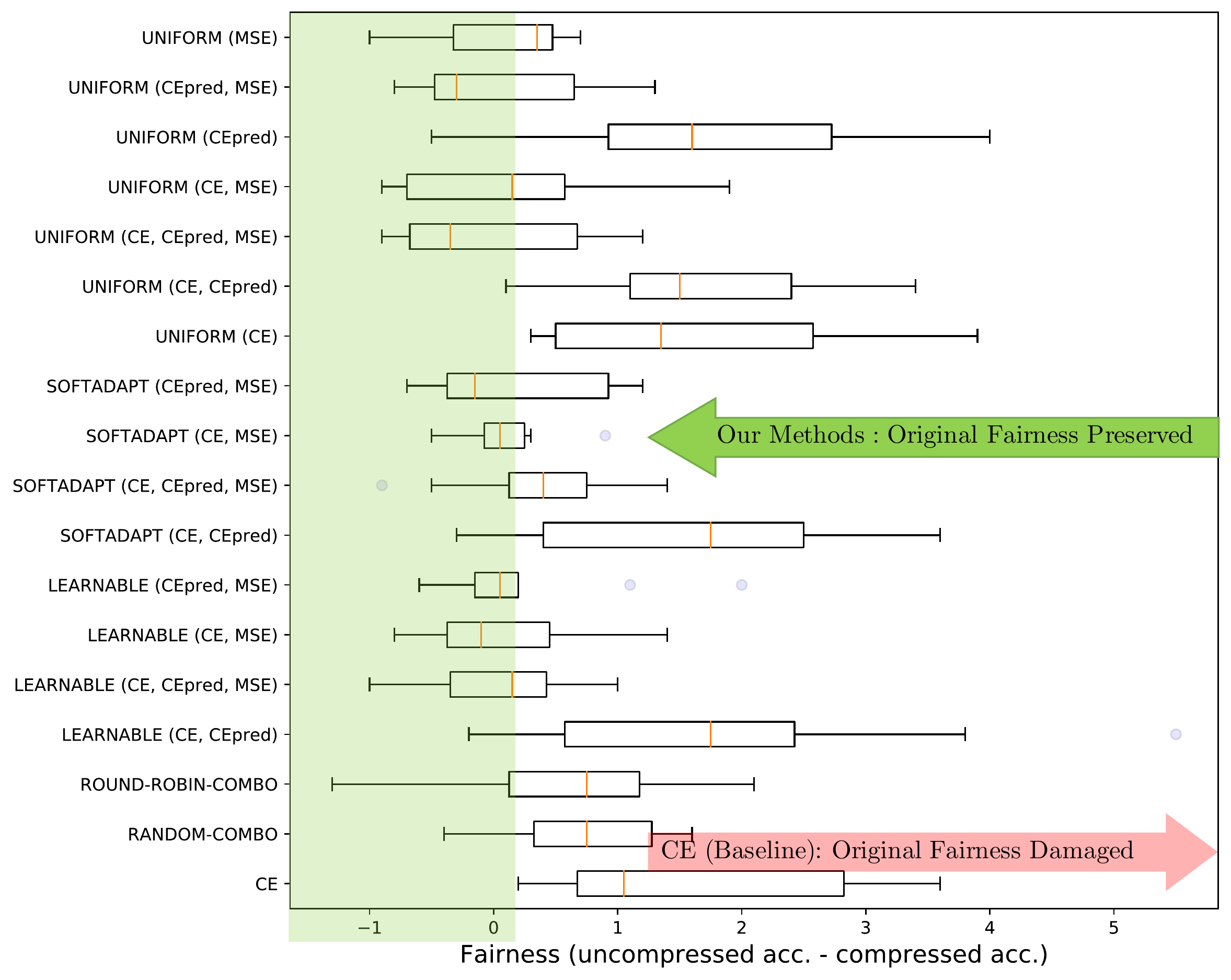}
    \label{fig:fairness_results_box}
    \end{subfigure}
    \vspace{-4mm}
    \caption{Effect of using our loss formulation on fairness. Left: Unfairness Function is computed as the difference between the maximum and the minimum class accuracies, among the different classes in this task. The wider the gap, the more unfair the class treatment. This is just one of many valid fairness measures, but it is particularly relevant for resource allocation in systems that tend to a ‘winner-takes-all’ state Paganini et al.~\cite{paganini2020prune}, (max\_acc - min\_acc). Right: a relative notion of fairness which plots the distributional shift between the accuracy of the uncompressed and the compressed models assuming the uncompressed model to fairness reference. Here, the blue points indicate outliers.}
    \label{fig:fairness_results}
\end{figure*}

We use the metrics mentioned in Section~\ref{sec:method} to compute the fairness of the model in Figure~\ref{fig:fairness_results}.
The plot on the left visualizes fairness (or rather, unfairness), computed as the difference between the maximum
and the minimum class accuracies; here, the wider the gap, the more unfair the class treatment. The plot on the right visualizes the distributional shift between the accuracies of the uncompressed and compressed models. As both the figures demonstrate, loss functions that include the logit pairing term continue to perform well; in particular, \textsc{SoftAdapt}-(CE, MSE), \textsc{Uniform}-(CEPred, MSE) and \textsc{Uniform}-(CE, CEPred, MSE) manage to largely preserve fairness despite compression.

\subsection{Attribution Preservation}\label{sec:attribute-iou}
\label{sec:attribution}

\begin{figure*}[t]
    \centering
    \begin{subfigure}[b]{0.45\textwidth}
    \includegraphics[width=\textwidth]{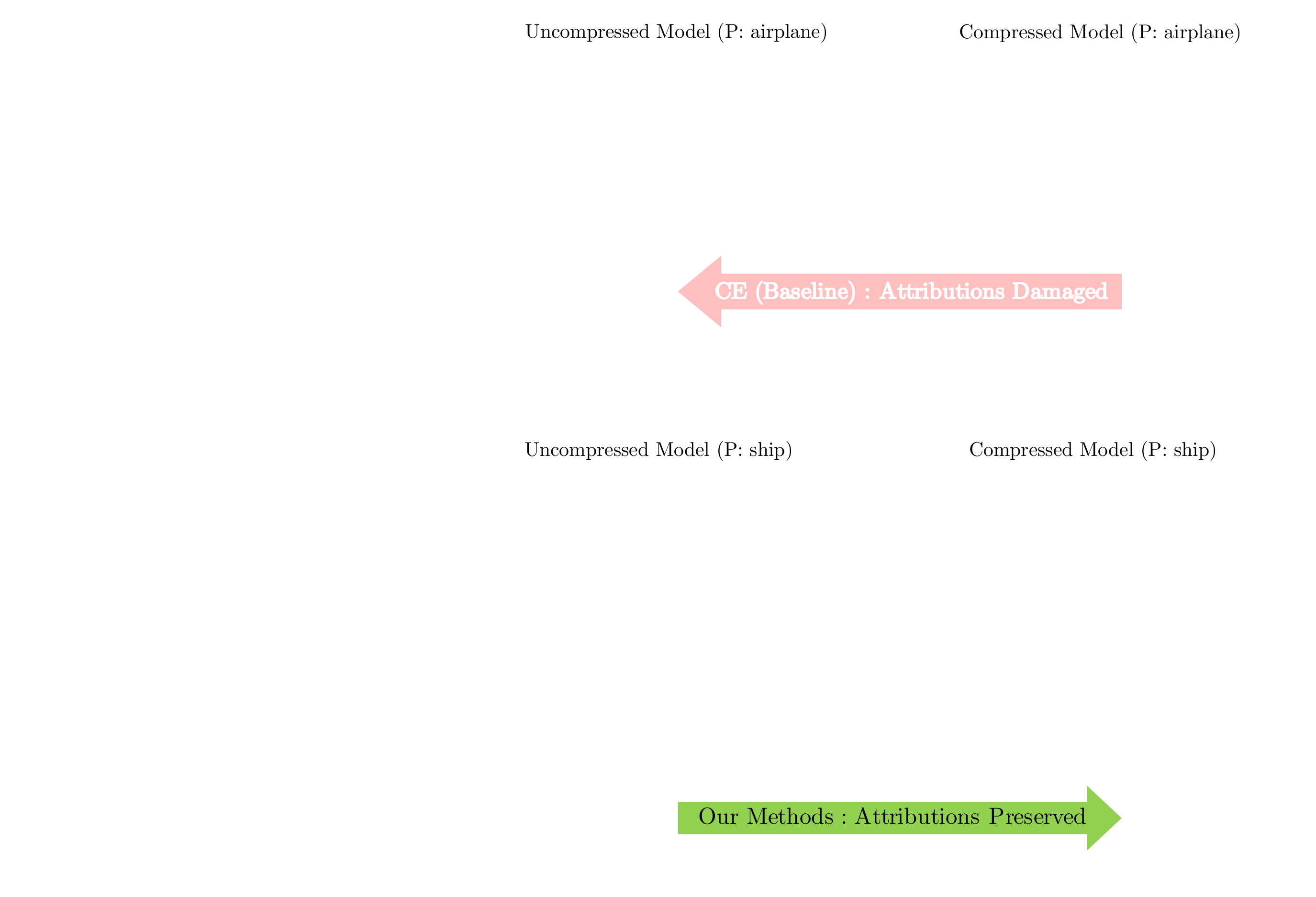}
    \end{subfigure}
    \begin{subfigure}[b]{0.38\textwidth}
    \includegraphics[width=\textwidth]{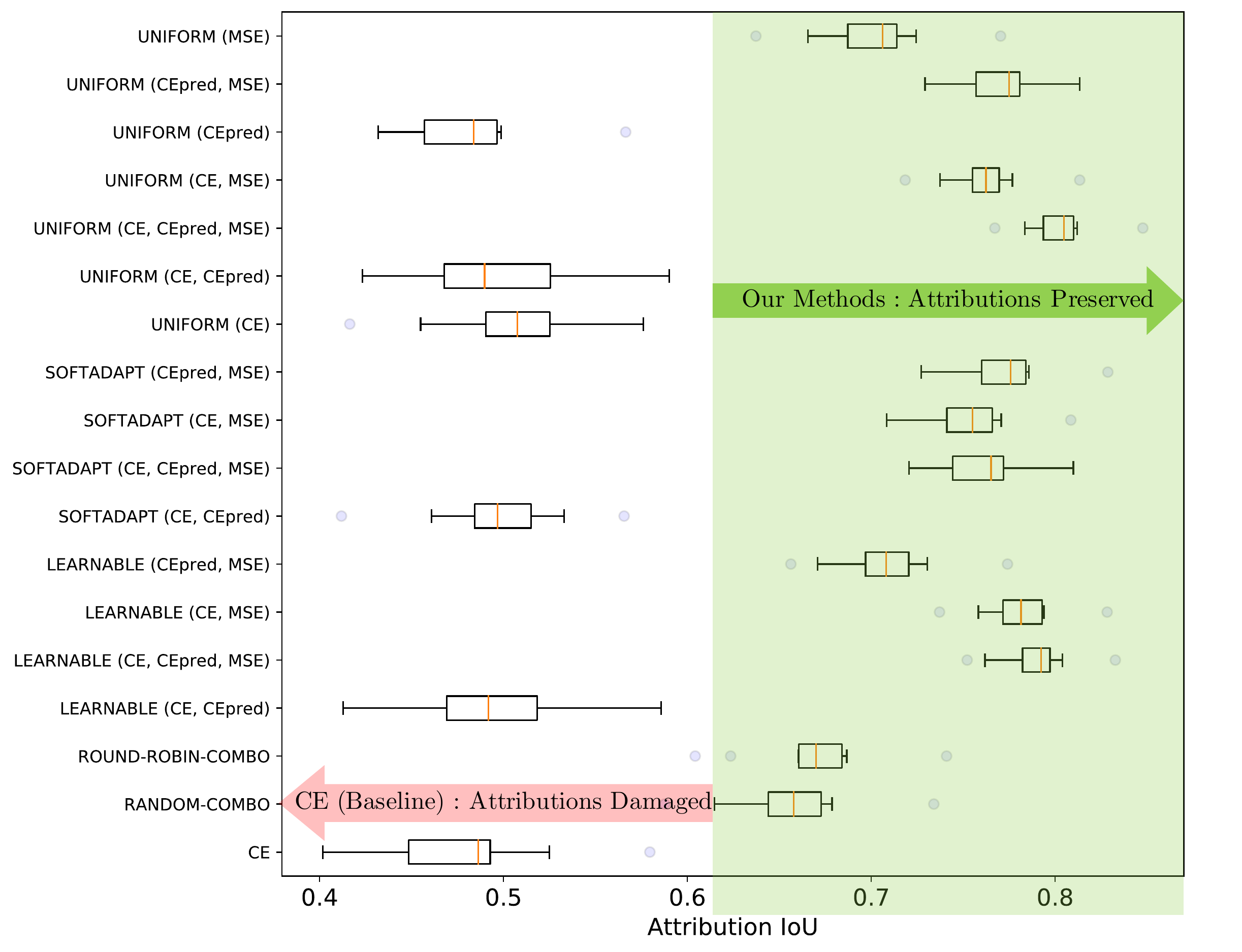}
    \end{subfigure}
    \caption{Attribution preservation results. Left: IoU between attributions on CIFAR-10. Here, the first column shows the original images, the second column visualizes the attributions from the uncompressed model while the third column visualizes the attributions from the compressed model. $T$ represents the target, while $P$ represents the prediction. Right: Attribution IoU for ResNet-20 (CIFAR-10).}
    \label{fig:attrib_iou}
\end{figure*}

As described in Section~\ref{sec:method}, we compute the IoU between the attributions of the compressed and the uncompressed models (ResNet-20 trained on CIFAR-10).
The distribution of attribution IoU for different classes is visualized in Figure~\ref{fig:attrib_iou} (Right) for ResNet-20 (CIFAR-10). As per our hypothesis, logit pairing in general gives a boost in attribution IoU, indicating the prediction alignment also has implications for attribution alignment. 
Figure~\ref{fig:attrib_iou} (Left) visualizes two examples along the corresponding attributions from the compressed and the uncompressed model where the first one achieves a low IoU while the second one achieves a high IoU.

\section{Conclusions}


This paper has presented a novel method for identifying and reducing the misalignment with the reference model during model compression. We study different loss functions inspired by the knowledge distillation literature and develop ways to combine them without extensive hyperparameter tuning. We also introduce metrics that capture {\em semantic} alignment between models. Using extensive experiments spanning different tasks, model architectures, we show that our methods lead to a significant improvement in these metrics. For metrics such as the number of  compression impacted exemplars (CIE), our methods help us obtain up to a $4.1\times$ reduction in CIEs compared to state-of-the-art benchmarks.  Our work demonstrates that distillation objectives may be useful as a general tool for preserving semantic similarity in model compression. 

\section*{Broader Impact}

As machine learning models are deployed in more and more edge devices, model compression has been taking a center stage in ensuring efficient and low-power inference. With models being deployed in devices like speakers and security cameras, guarantees on class-specific accuracy and feature-similarity with the reference model become important design goals. Our work suggests metrics that can be used broadly in such applications. This said, we highlight that the list of metrics we consider is not exhaustive and concerns such as privacy will need other metrics. 

Ultimately, effective model compression will also allow power-efficient deployment of ML methods, which will likely have a positive environmental impact. 

\section*{Acknowledgements}

This work is in part supported by NSF Awards 1704715 and 1817073. This work was in part also supported by the BMBF project DeFuseNN (Grant 01IW17002) and the NVIDIA AI Lab (NVAIL) program. Vinu Joseph is supported in part by an NVIDIA Graduate Research Fellowship. AB acknowledges the support from NSF CCF-2008688.

\bibliographystyle{plain}
\bibliography{paper}

\clearpage

\appendix

\section{Details on Hyperparameter Learning}\label{sec:app:learnable}

Section~\ref{sec:method} describes three methods for tuning the parameters $\alpha, \beta$, and $\gamma$ in our combined loss objective. We now provide the details of \textsc{Learnable} and \textsc{SoftAdapt} procedures.

In the \textsc{Learnable} variant, as described in the section, we need to prevent the objective from losing dependence on some of the terms. We achieve this via a weight decay term. The details are as follows. We set
\begin{equation*}
    \alpha = \frac{e^{\alpha'}}{e^{\alpha' + \beta' + \gamma'}},
\end{equation*}
and define $\beta$ and $\gamma$ similarly. Here $\alpha'$ are parameters of the model.  Once the final loss value is computed based on these weights, the parameters of the model (including $\alpha'$, $\beta'$ and $\gamma'$) are updated using SGD.

\begin{equation*}
    \alpha' = \alpha' - \nabla_{\alpha'} (\mathcal{L} + \eta \norm{\alpha'}^2) \\
\end{equation*}

\noindent where $\eta$ represents the weight decay. We use a strong weight-decay of 1.0 in our experiments to ensure that a particular loss strictly does not dominate others.

For {\text{SoftAdapt}}, the details are as follows. We have a parameter $\eta$ that controls the change in weights. Let $s_{\alpha} = \mathcal{L}_{\tce}(t) - \mathcal{L}_{\tce}(t-1)$ be the corresponding change in the loss value between two consecutive steps (define $s_\beta$ and $s_\gamma$ using the corresponding losses). 
They use the normalized version of SoftAdapt which can be written as:


\begin{equation*}
    s_{\alpha} = \frac{s_{\alpha}}{(s_{\alpha} + s_{\beta} + s_{\gamma}) + \epsilon},
\end{equation*}

\noindent where $\epsilon$ is introduced for numerical stability. The final weight based on these normalized scores is:

\begin{equation*}
    \alpha = \frac{e^{\eta s_{\alpha}}}{e^{\eta s_{\alpha}} + e^{\eta s_{\beta}} + e^{\eta s_{\gamma}}},
\end{equation*}

where $\eta$ selects whether to optimize the worst or the best loss based on whether the value of $\eta$ is greater than or less than zero respectively. We use $\eta = 1.0$ in our experiments which equates to optimizing the worst performing loss at every weight update step. Note that these weights are not optimized at every step of the optimization process, but updated after every 10 optimization steps and the corresponding average change is taken into account.

\section{Method Overview}

\begin{figure*}[t]
\centering
  \includegraphics[width=0.85\textwidth]{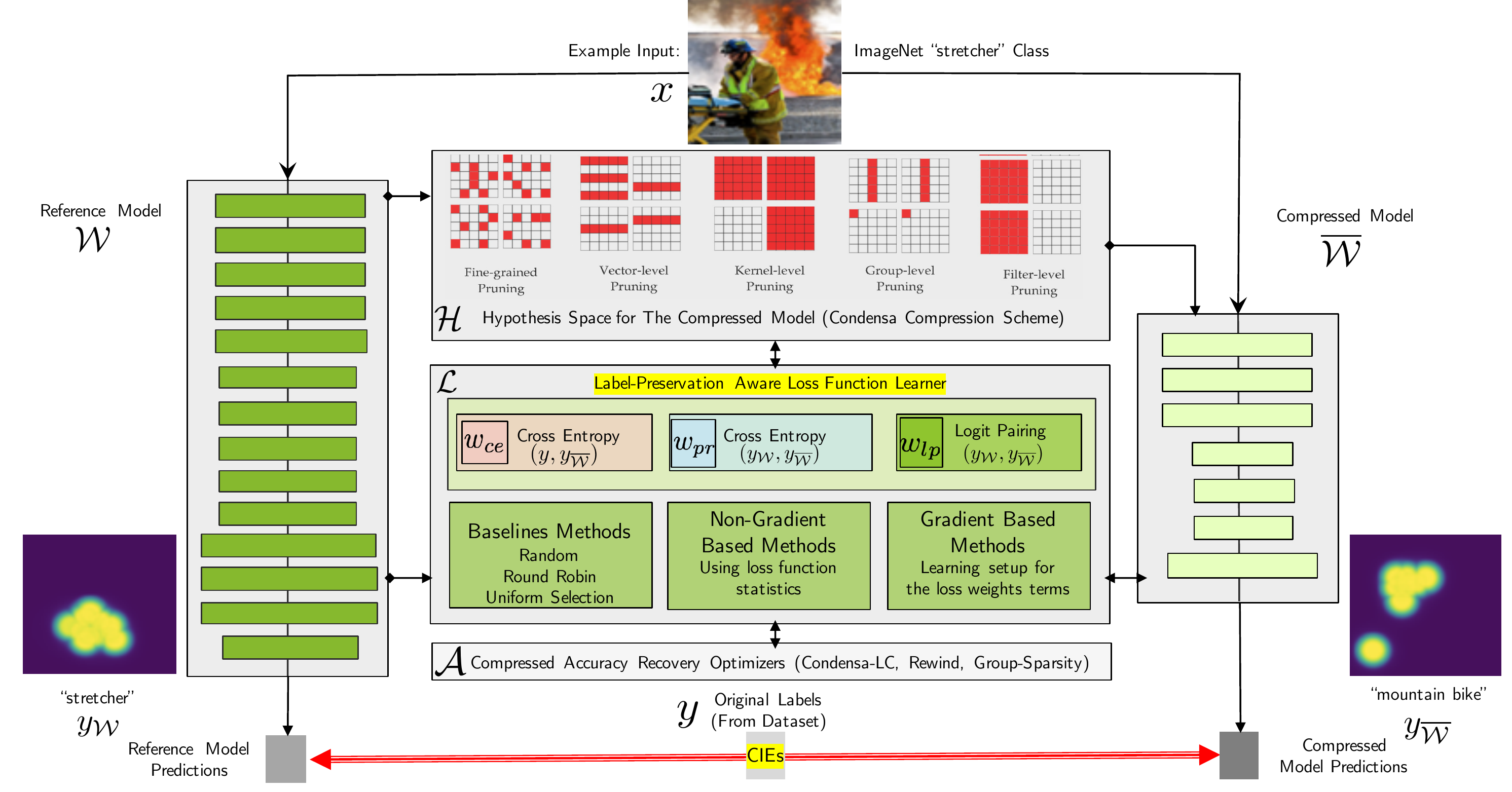}
  \caption{Overview of the experimental setup used for evaluating various label preservation-aware loss functions.}
  \label{fig:overview}
\end{figure*}

The overview of our framework is visualized in Fig.~\ref{fig:overview}.
It can be seen that the misalignment between the compressed and the uncompressed models can be evaluated at several different levels, starting from predictions, to attribution maps.
The figure also presents an overview of our label-preservation-aware loss function which is comprised of three different terms (CE, MSE and CE\_pred) whose individual contributions can be defined using different weighting schemes such as uniform weighting, soft-adapt, or a learnable weighting scheme based on the gradients.

\section{Ramifications of Model Compression}

\begin{figure}[t]
    \centering
    \includegraphics[width=1.0\linewidth]{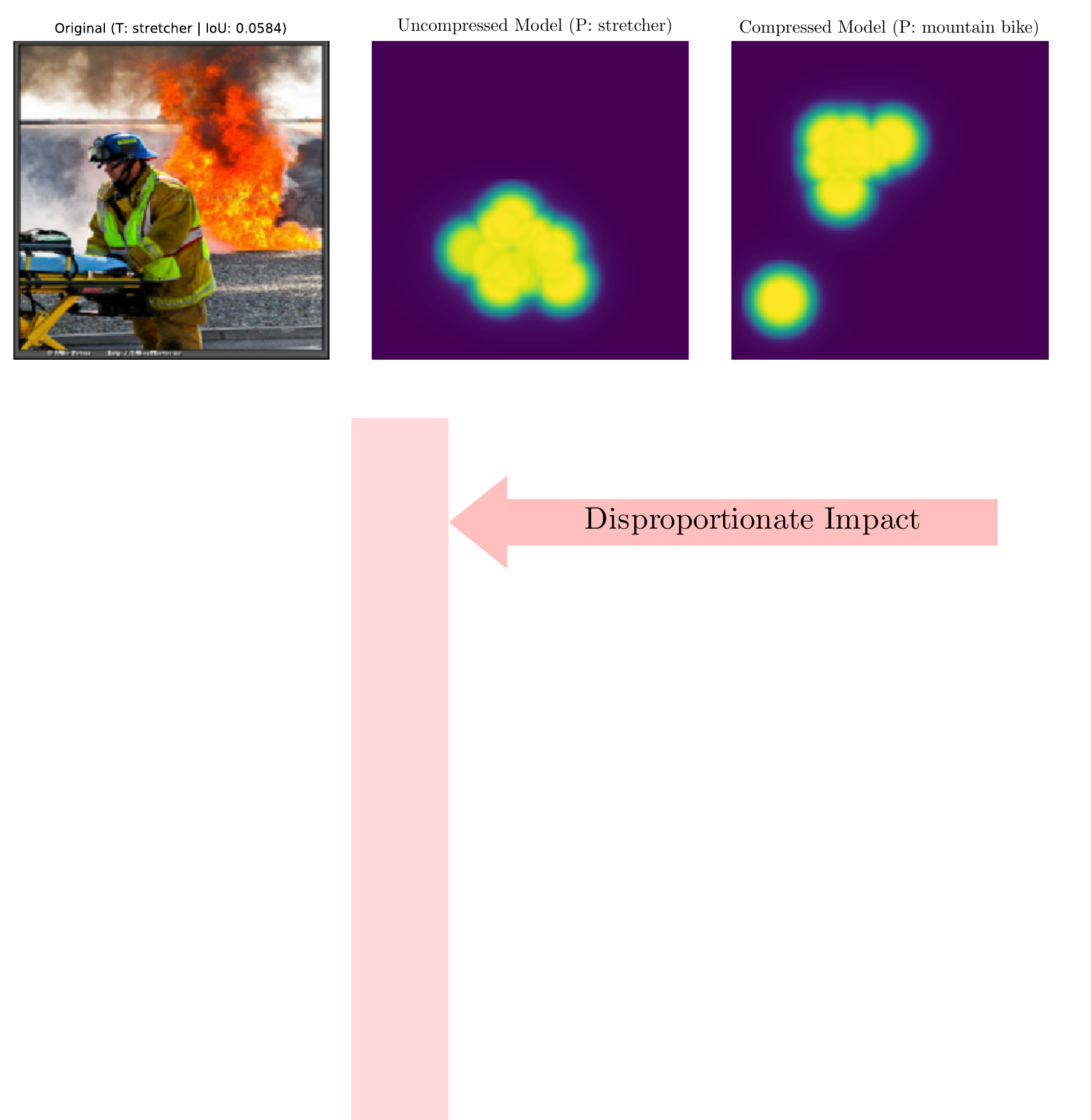}
    \caption{\small Top: Attribution mismatch for reference and compressed models; Bottom: Disparate class-level impact on accuracy after model compression. Note the significant difference in the predictions of the compressed and the uncompressed models as quantified by the number of CIEs.}
    \label{fig:intro}
\end{figure}

Fig.~\ref{fig:intro} represents two ramifications of model compression.
The first ramification is related to features where the compressed and the uncompressed models focus on an entirely different set of features. This has been discussed in detail in Section~\ref{sec:intro}.
The second ramification is related to desperate impact over classes, where some of the classes are more affected than others are model compression.
These fairness ramifications are discussed in Section~\ref{sec:fairness}.

\section{Accuracy Recovery Algorithms in Model Compression Experiments}

Deep neural networks are heavy on computation and memory by design,
creating an impediment to operating these networks on resource-constrained platforms.
To alleviate this constraint, several branches of work 
have been proposed to reduce the size of an existing
neural network.
The most commonly employed approach 
is to reduce the number of weights, neurons, or layers in a 
network while maintaining approximately the same
performance~\cite{joseph2020programmable}.
This approach was first explored on DNNs
in early work such as~\cite{hassibi1994optimal,lecun1990optimal}.%
Studies conducted by~\cite{han2015deep,han2015learning} 
showed that simple unstructured pruning can reduce the size
of the network by pruning unimportant connections within the 
network. However, such unstructured pruning strategies produce large sparse weight matrices that are computationally inefficient unless equipped with a specialized hardware~\cite{numenta20}.
To resolve this issue, structured pruning methods were proposed 
where entire channels are pruned simultaneously to ensure that the pruned network can be naturally accelerated on commodity hardware~\cite{wen2016learning,li2016pruning,hu2016network}. 
More recently, Renda et al.~\cite{renda2020rewind} proposed the \textit{rewind} algorithm which is similar to simple fine-tuning of the network to regain the loss in accuracy incurred during the pruning step. The sparsity level of the model is updated in small steps where each step enhances the sparsity of the model followed by fine-tuning.
The two major schemes for structured pruning are either based on filter pruning~\cite{joseph2019condensa} or low-rank tensor factorization~\cite{li2020group}. Both these approaches enable direct acceleration of the networks in contrast to unstructured pruning.
Li et al.~\cite{li2020group} explored the relationship between tensor factorization and general pruning methods, and proposed a unified approach based on sparsity-inducing norm which can be interpreted as both tensor factorization or direct filter pruning. By simply changing the way the sparsity regularization is enforced, filter pruning and low-rank decomposition can be derived accordingly. 
This is particularly important for the compression of popular network architectures with shortcut connections (e.g. ResNet), where filter pruning cannot deal with the last convolutional layer in a ResBlock due to change in the output dimensions while the low-rank decomposition methods can.

\subsection{Accuracy Recovery Algorithms} General accuracy recovery algorithms capable of handling a wide variety of compression techniques provide the foundation for modern compression systems. Prior work in this domain includes the LC algorithm~\cite{carreira2017model}, ADAM-ADMM~\cite{zhang2018adam} and DCP~\cite{zhuang2018discrimination}. More recently,
the \textit{Rewind}~\cite{renda2020rewind} and Group-Sparsity~\cite{li2020group} algorithms have been demonstrated to be state-of-the-art compression algorithms.
Due to their compression scheme-agnostic nature, we build upon these two methods in our paper to evaluate the proposed \emph{label-preservation-aware loss functions} as described in Section~\ref{sec:minimize_misalignment}.
%

\subsection{Network Distillation}
Another branch of network compression initially proposed by \cite{hinton2015distilling}, attempts to distill knowledge from a large teacher network to a small student network. 
%
With the assumption that the knowledge captured by a network is reflected in the output probability distribution, this line of work trains the student network to mimic the probability distribution produced by the teacher network. Since the networks are trained to output one-hot distribution, a temperature $T$ is used to diffuse the probability mass.
Advanced methods of distillation have succeeded in achieving much more effective transfer by not only
transferring the output logits but the information of the intermediate activations as in~\cite{ahn2019variational, zagoruyko2016paying, romero2014fitnets, jang2019learning}.
Although network distillation was presented as a general form of logit pairing, it is quite difficult to obtain improvements during distillation without spending considerable effort in manually tuning the temperature $T$ for the softmax layer. In contrast, using pure logit pairing comes without any additional cost of manual hyperparameter tuning.
Therefore, we employ pure logit pairing instead of knowledge distillation in our approach.

%
%

\subsection{Group-Sparsity based Model Compression}

We now briefly describe the key insight of the compression recovery algorithm that was used in our evaluation.
The main idea in the Group-Sparsity recovery algorithm \cite{li2020group} is that the filter pruning and filter decomposition seek a compact
approximation of the parameter tensors despite their different operational forms to cope with different application scenarios.
Consider a vectorized image patch  $\bf{x} \in \mathbb{R}^{m \times 1}$
and a group of $n$ filters $\bf{\mathcal{W}} = \{\bf{w_1}, \cdots , \bf{w_n}\} \in \mathbb{R}^{m \times n}$.
The pruning methods remove output channels and approximate the original output $\bf{x}^T \bf{\mathcal{W}}$ as $\bf{x}^T \bf{C}$, where $\bf{C} \in \mathbb{R}^{m \times k}$ only has
$k$ output channels. Filter decomposition methods approximate $\bf{\mathcal{W}}$ as two filters $\bf{A} \in \mathbb{R}^{m\times k}$
and $\bf{B} \in \mathbb{R}^{k \times n}$, making $\bf{AB}$
a rank $k$ approximation of $\bf{\mathcal{W}}$. 
Thus, both pruning and decomposition-based methods seek a compact approximation to the original network parameters, but adopt different strategies for the approximation.
The weight parameters $\bf{\mathcal{W}}$ are usually trained with some regularization 
such as weight decay to constrain the hypothesis class.
To get structured pruning of the filter, structured sparsity regularization 
is used to constrain the filter:
\begin{equation}
\min _{\mathcal{W}} \mathcal{L}(y, \Phi(\mathbf{x} ; \mathcal{W}))+\mu \mathcal{D}(\mathcal{W})+\lambda \mathcal{R}(\mathcal{W})
\label{eq:loss1}
\end{equation}
where $\mathcal{D}(\cdot)$ and $\mathcal{R}(\cdot)$ represents the weight decay and 
sparsity regularization term respectively, while $\mu$ and $\lambda$ are the regularization factors.
Instead of directly regularizing the matrix $\bf{\mathcal{W}}$ \cite{yoon2017combined, li2019oicsr}, we enforced group sparsity constraints by incorporating a sparsity-inducing matrix $\mathbf{A} \in \mathbb{R}^{n \times n}$, which can be converted to the filter of a $1 \times 1$ convolution layer after the original layer.
Then the original convolution of 
$Z = X \times \mathcal{W}$ becomes 
$Z = X \times (\mathcal{W} \times \mathbf{A})$.
To obtain a structured sparse matrix, group sparsity regularization is enforced on $\bf{A}$. Thus, the loss in Eqn. \ref{eq:loss1} function becomes
\begin{equation}
\min _{\mathcal{W}, \mathbf{A}} \mathcal{L}(y, \Phi(\mathbf{x} ; \mathcal{W}, \mathbf{A}))+\mu \mathcal{D}(\mathcal{W})+\lambda \mathcal{R}(\mathbf{A})
\label{eq:loss2}
\end{equation}
Solving the problem in Eqn. \ref{eq:loss2} results in structured group
sparsity in matrix $\textbf{A}$. By considering matrix $\bf{\mathcal{W}}$ and $\bf{A}$ together, the actual effect is that the original convolutional
filter is compressed.


\section{Application to Model Compression in Sensitive Domains (Brain MRI FLAIR Segmentation)}

\begin{figure*}[t]
    \centering
    
    \begin{subfigure}[b]{0.15\textwidth}
        \includegraphics[width=\textwidth]{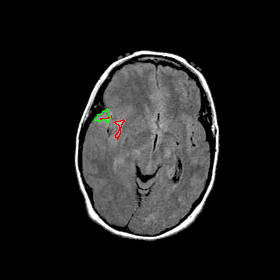}
    \end{subfigure}
    \begin{subfigure}[b]{0.15\textwidth}
        \includegraphics[width=\textwidth]{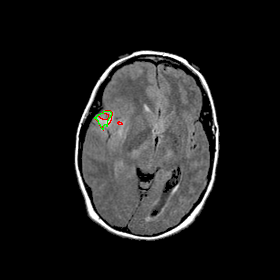}
    \end{subfigure}
    \begin{subfigure}[b]{0.15\textwidth}
        \includegraphics[width=\textwidth]{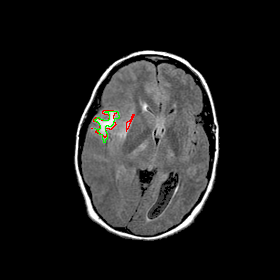}
    \end{subfigure}
    \begin{subfigure}[b]{0.15\textwidth}
        \includegraphics[width=\textwidth]{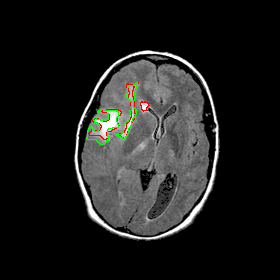}
    \end{subfigure}
    \begin{subfigure}[b]{0.15\textwidth}
        \includegraphics[width=\textwidth]{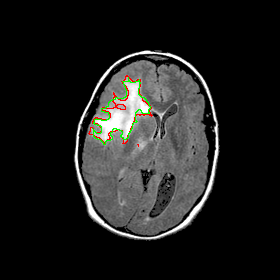}
    \end{subfigure}
    \begin{subfigure}[b]{0.15\textwidth}
        \includegraphics[width=\textwidth]{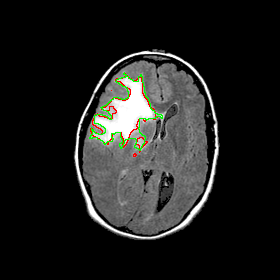}
    \end{subfigure}

    \begin{subfigure}[b]{0.15\textwidth}
        \includegraphics[width=\textwidth]{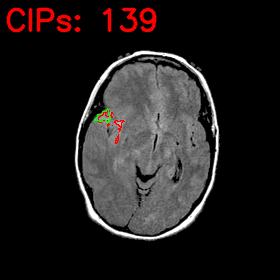}
    \end{subfigure}
    \begin{subfigure}[b]{0.15\textwidth}
        \includegraphics[width=\textwidth]{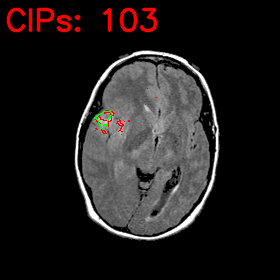}
    \end{subfigure}
    \begin{subfigure}[b]{0.15\textwidth}
        \includegraphics[width=\textwidth]{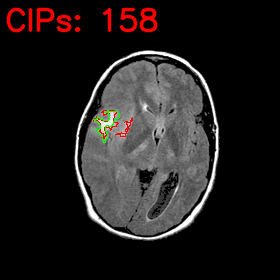}
    \end{subfigure}
    \begin{subfigure}[b]{0.15\textwidth}
        \includegraphics[width=\textwidth]{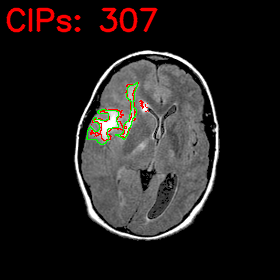}
    \end{subfigure}
    \begin{subfigure}[b]{0.15\textwidth}
        \includegraphics[width=\textwidth]{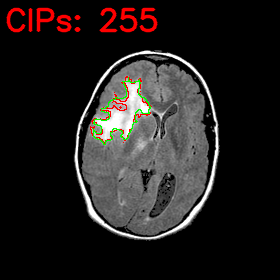}
    \end{subfigure}
    \begin{subfigure}[b]{0.15\textwidth}
        \includegraphics[width=\textwidth]{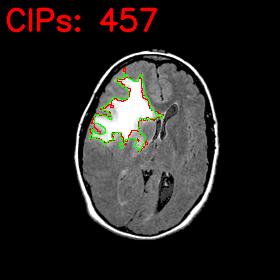}
    \end{subfigure}

    \begin{subfigure}[b]{0.15\textwidth}
        \includegraphics[width=\textwidth]{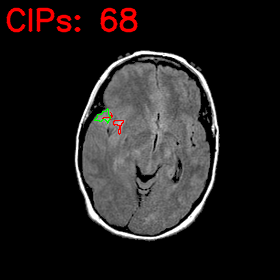}
    \end{subfigure}
    \begin{subfigure}[b]{0.15\textwidth}
        \includegraphics[width=\textwidth]{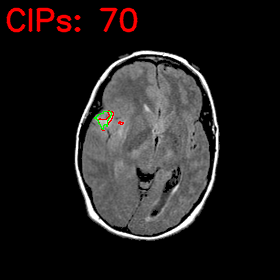}
    \end{subfigure}
    \begin{subfigure}[b]{0.15\textwidth}
        \includegraphics[width=\textwidth]{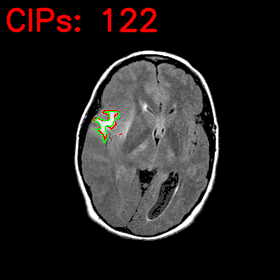}
    \end{subfigure}
    \begin{subfigure}[b]{0.15\textwidth}
        \includegraphics[width=\textwidth]{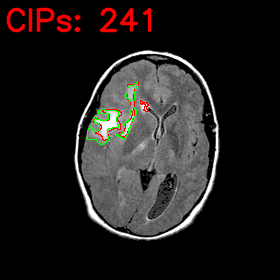}
    \end{subfigure}
    \begin{subfigure}[b]{0.15\textwidth}
        \includegraphics[width=\textwidth]{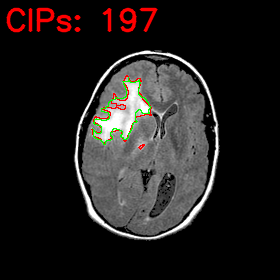}
    \end{subfigure}
    \begin{subfigure}[b]{0.15\textwidth}
        \includegraphics[width=\textwidth]{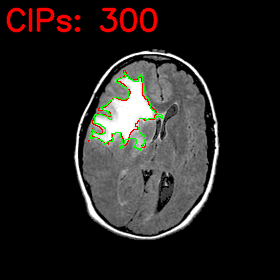}
    \end{subfigure}

    \begin{subfigure}[b]{0.15\textwidth}
        \includegraphics[width=\textwidth]{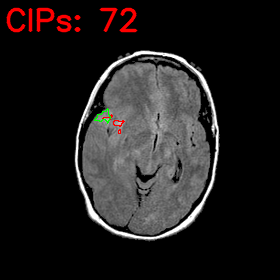}
    \end{subfigure}
    \begin{subfigure}[b]{0.15\textwidth}
        \includegraphics[width=\textwidth]{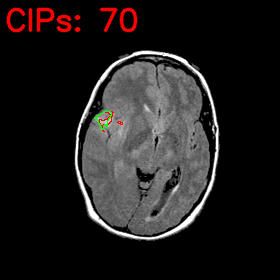}
    \end{subfigure}
    \begin{subfigure}[b]{0.15\textwidth}
        \includegraphics[width=\textwidth]{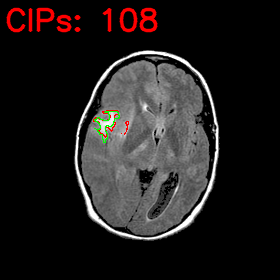}
    \end{subfigure}
    \begin{subfigure}[b]{0.15\textwidth}
        \includegraphics[width=\textwidth]{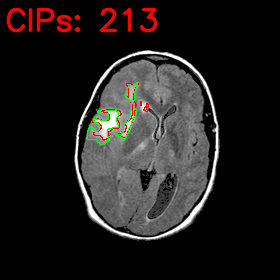}
    \end{subfigure}
    \begin{subfigure}[b]{0.15\textwidth}
        \includegraphics[width=\textwidth]{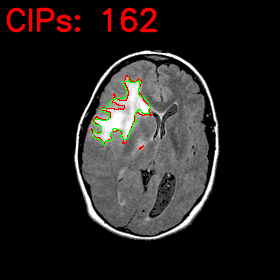}
    \end{subfigure}
    \begin{subfigure}[b]{0.15\textwidth}
        \includegraphics[width=\textwidth]{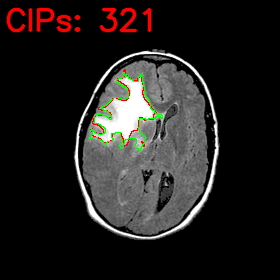}
    \end{subfigure}

    \caption{CIP reduction on a small slice of Brain MRI images. 
    First row represents the uncompressed model, second row represents the model compressed using Uniform (CE), third row represents the model compressed using Uniform (MSE), while the fourth row represents the model compressed using Uniform (CE, MSE).
    Red contour represents the prediction while the green contour represents the ground-truth.}
    \label{fig:brain_mri_seg}
\end{figure*}

Fig.~\ref{fig:brain_mri_seg} presents a qualitative assessment of the segmentation masks produced by different compression techniques for the FLAIR Brain MRI Segmentation task.
It can see that the number of CIPs are significantly lower when including the logit-pairing term (third row) as compared to only CE (second row).
It goes further down when including back the CE term as visualized in the last row.

\section{High Resolution Images}

\begin{figure*}[h]
    \centering
    \includegraphics[width=\textwidth]{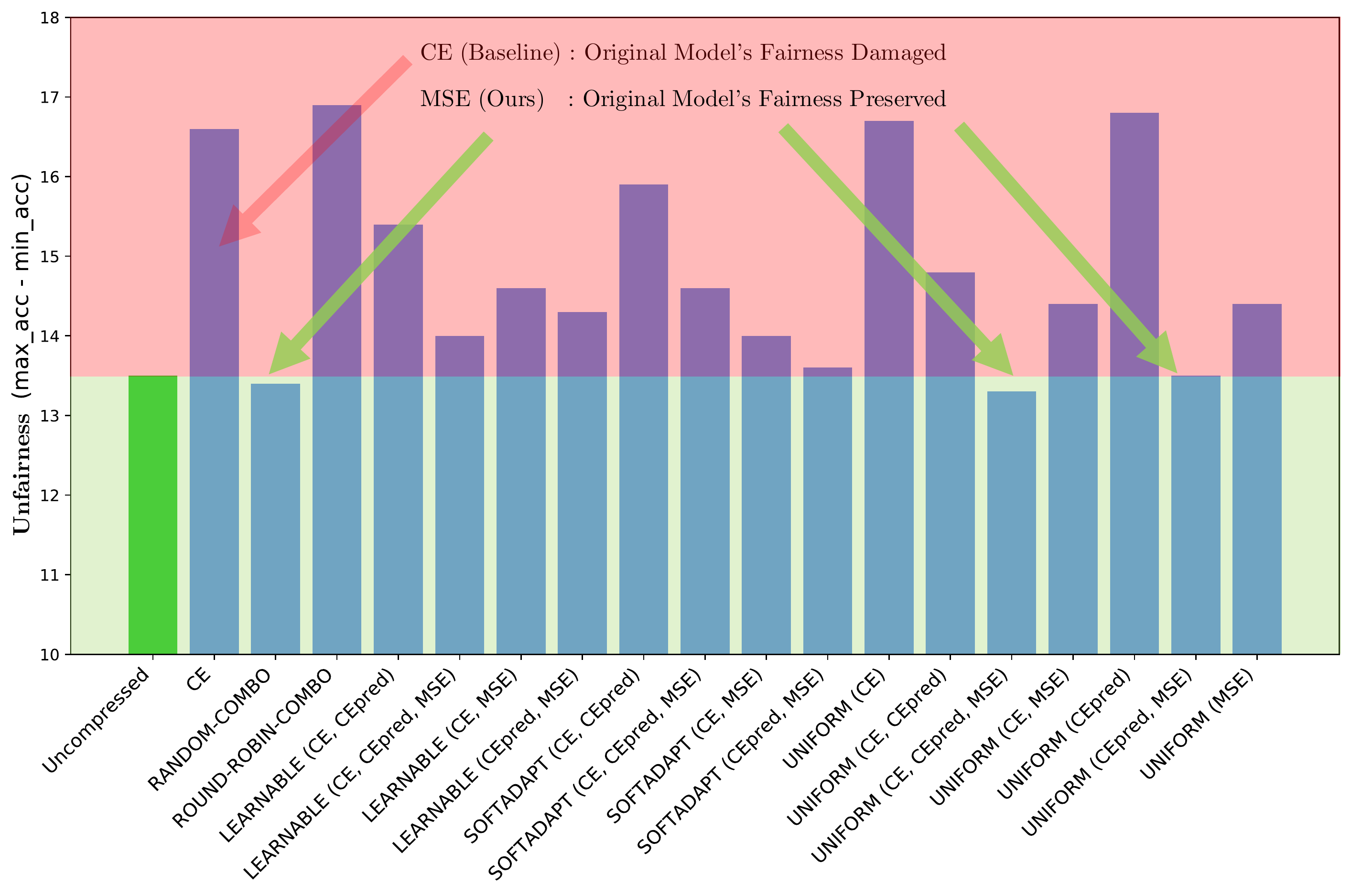}
    \caption{A higher resolution version of Fig.~\ref{fig:fairness_results} (left).}
    \label{fig:fairness_results_a}
\end{figure*}

\begin{figure*}[h]
    \centering
    \includegraphics[width=\textwidth]{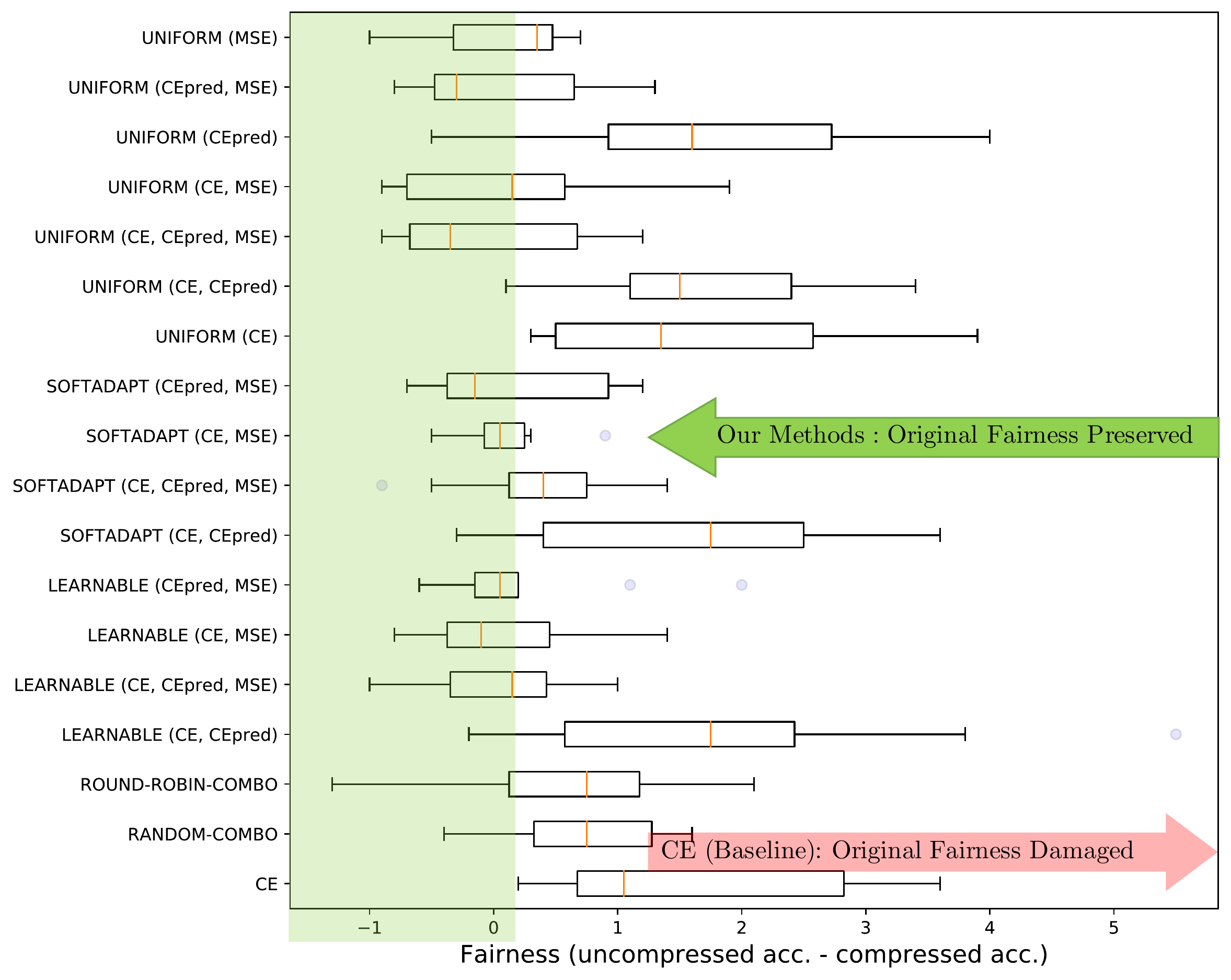}
    \caption{A higher resolution version of Fig.~\ref{fig:fairness_results} (right).}
    \label{fig:fairness_results_b}
\end{figure*}

\begin{figure*}[t]
    \centering
    \includegraphics[width=\textwidth]{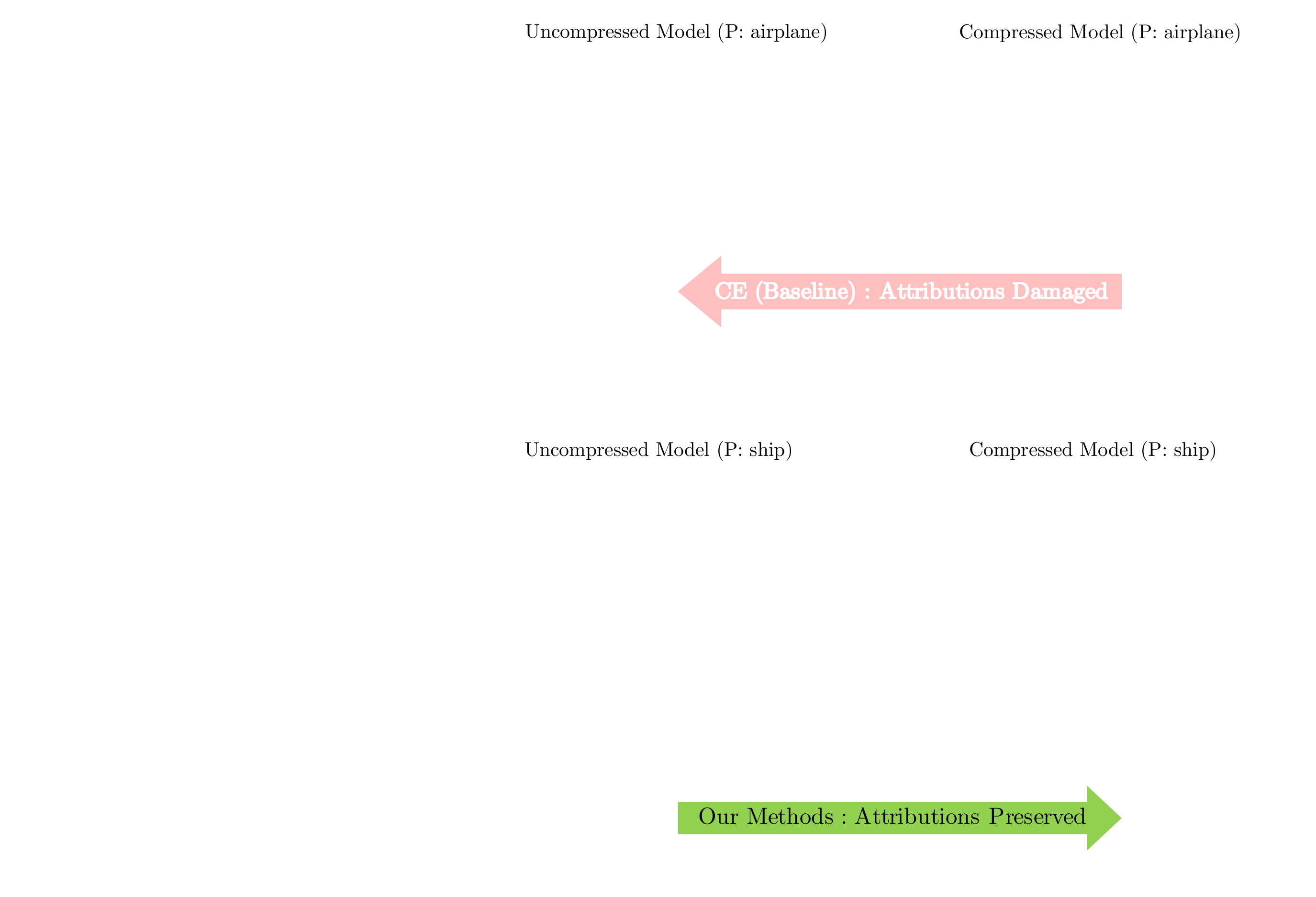}
    \caption{A higher resolution version of Fig~\ref{fig:attrib_iou} (left).}
    \label{fig:attrib_iou_a}
\end{figure*}

\begin{figure*}[t]
    \centering
    \includegraphics[width=\textwidth]{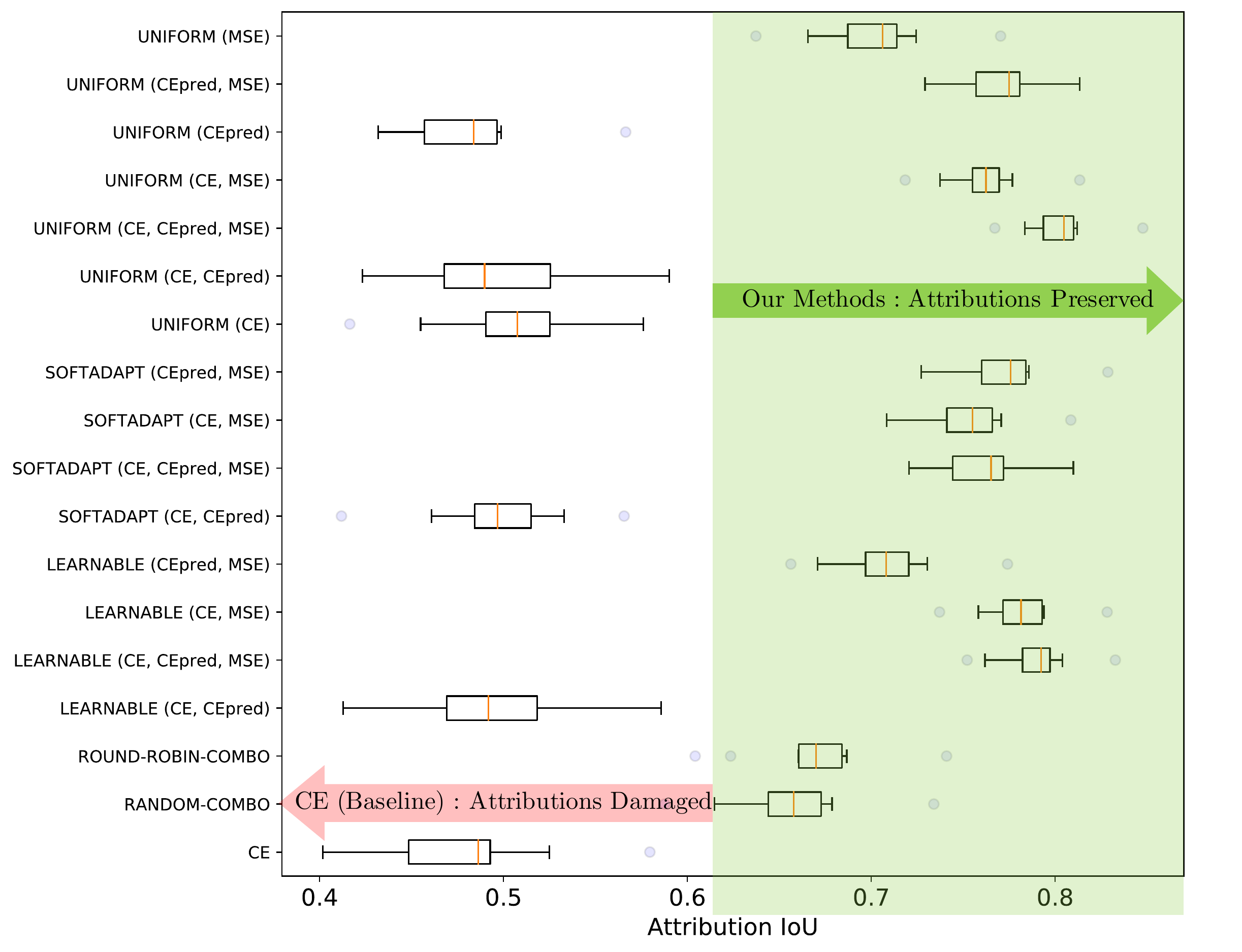}
    \caption{A higher resolution version of Fig~\ref{fig:attrib_iou} (right).}
    \label{fig:attrib_iou_b}
\end{figure*}

We split and reinclude Fig.~\ref{fig:attrib_iou} with a higher resolution for clarity in Fig.~\ref{fig:fairness_results_a} and Fig.~\ref{fig:fairness_results_b}.
Similarly, we also split and reinclude Fig.~\ref{fig:attrib_iou} in Fig.~\ref{fig:attrib_iou_a} and Fig.~\ref{fig:attrib_iou_b}.
This helps in detailed inspection of results, which was constrained due to space limitations in the main body of the paper.

\FloatBarrier

\section{Additional Experimental Results}

We now present the results for all the evaluated models on CIFAR-10 and CIFAR-100 with all the evaluated loss combinations.
All the results are consistent with the primary results presented in the paper on ImageNet (Fig.~\ref{fig:resnet50_imagenet}) where we see a significant reduction in CIEs when including the logit-pairing term, while simultaneously having a positive impact on top-1 accuracy.
This provides an in-depth overview of the consistency of the logit-pairing term across datasets and architectures for promoting model alignment.

\subsection{Results on CIFAR-10}

\subsubsection{DenseNet}

\begin{figure*}[!h]
    \centering
    \includegraphics[width=0.9\textwidth]{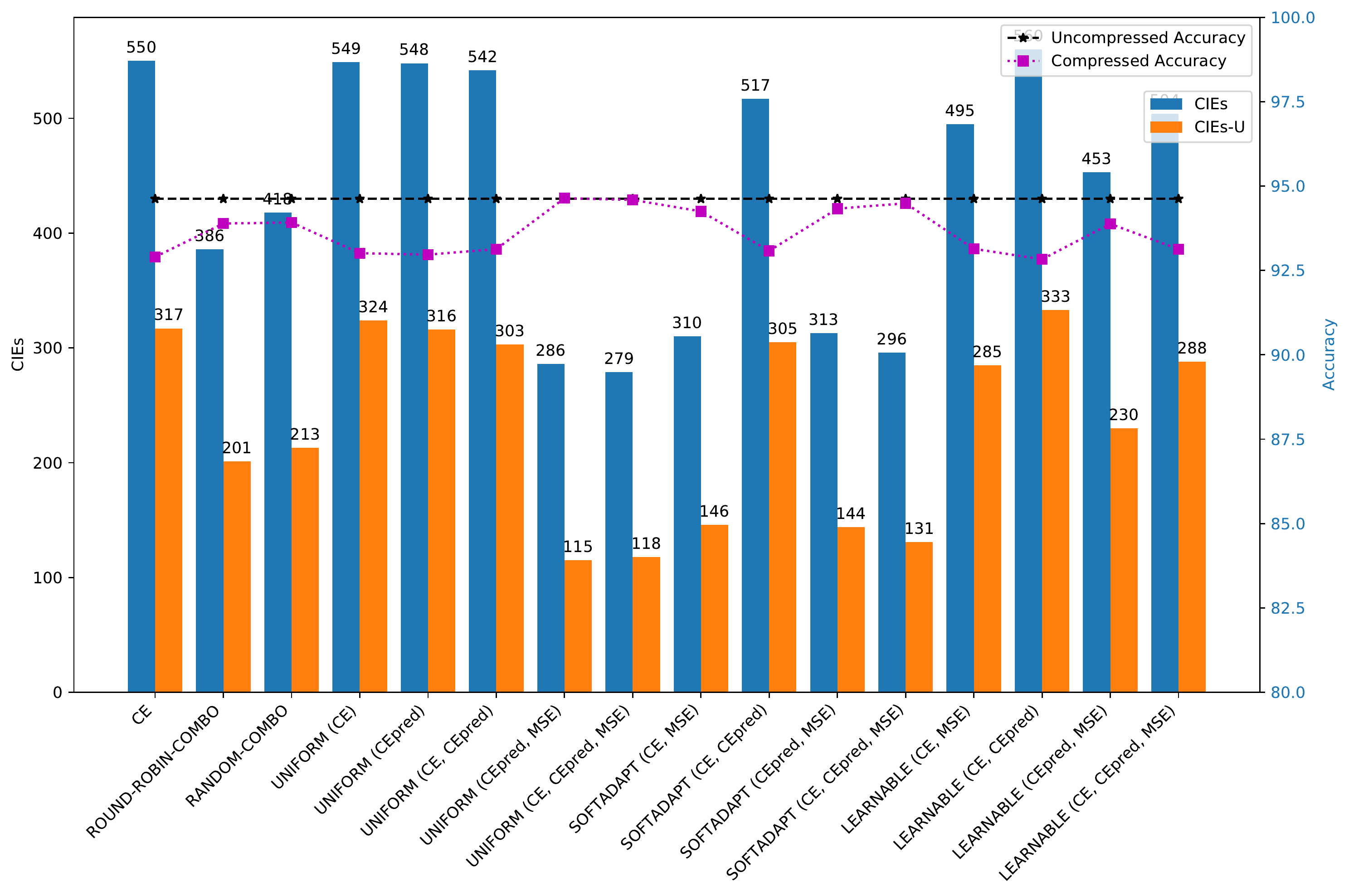}
    \caption{DenseNet results on CIFAR-10}
    \label{fig:densenet_cifar10_sup}
    \vspace*{-8mm}
\end{figure*}

The results for DenseNet on CIFAR-10 are presented in Fig.~\ref{fig:densenet_cifar10_sup}.

\subsubsection{ResNet-20}

\begin{figure*}[!h]
    \centering
    \includegraphics[width=0.9\textwidth]{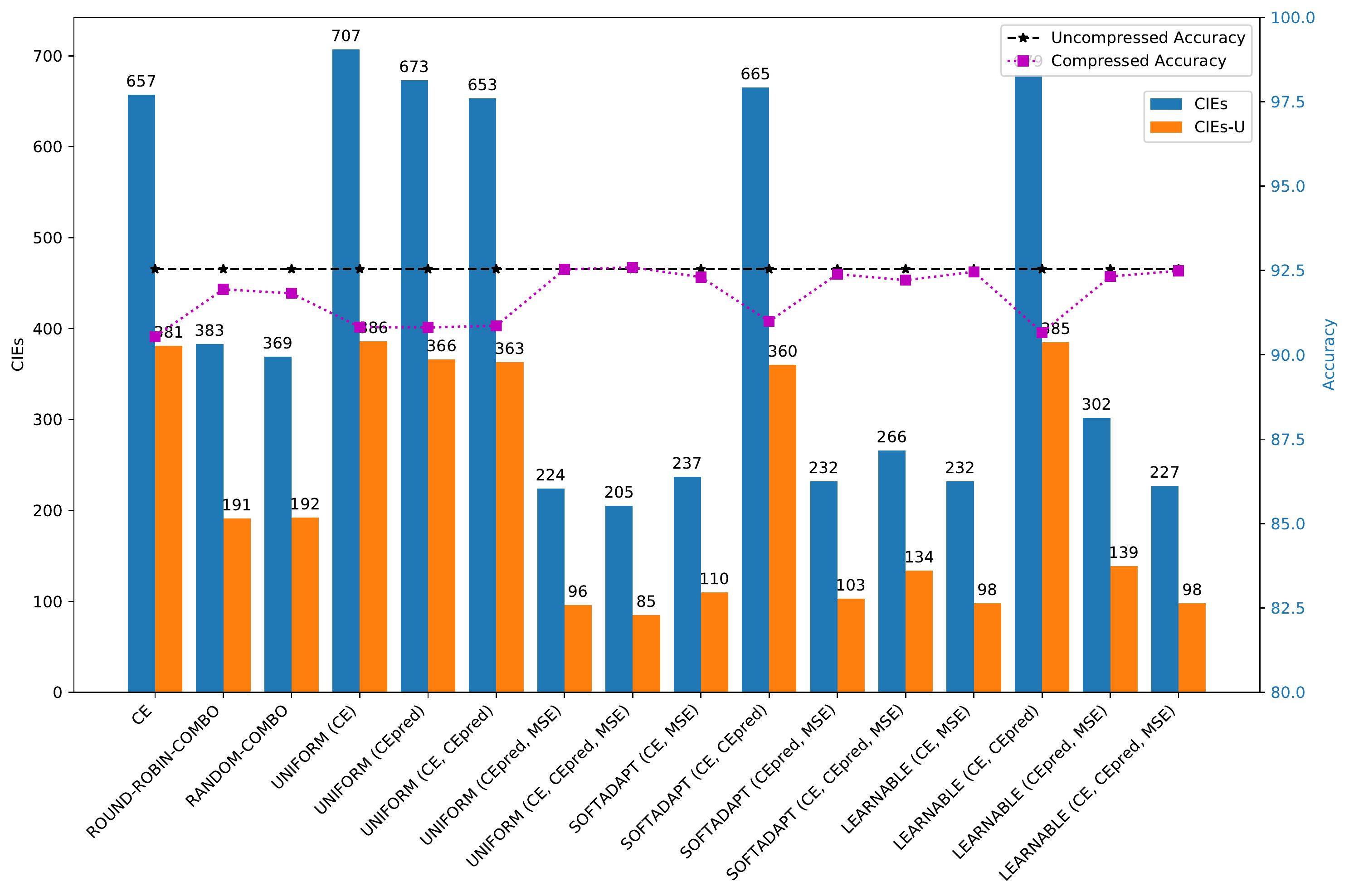}
    \caption{Performance of each loss+optimizer combination on ResNet-20 (CIFAR-10)}
    \label{fig:resnet20_cifar10_sup}
\end{figure*}
\clearpage

The results for ResNet-20 on CIFAR-10 are presented in Fig.~\ref{fig:resnet20_cifar10_sup}.

\subsubsection{ResNeXt-20}

\begin{figure*}[!h]
    \centering
    \includegraphics[width=0.9\textwidth]{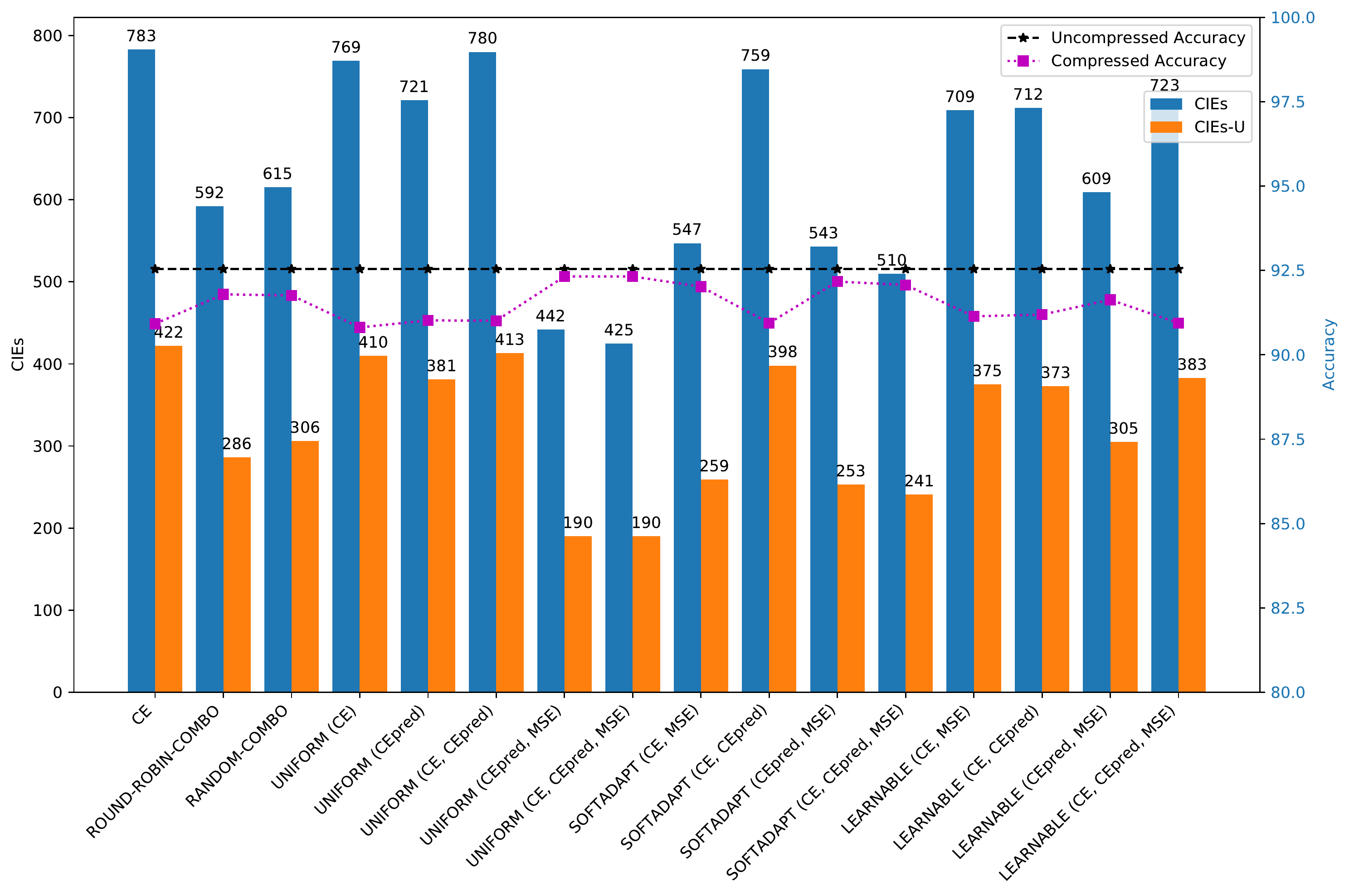}
    \caption{Performance of each loss+optimizer combination on ResNeXt-20 (CIFAR-10)}
    \label{fig:resnext20_cifar10_sup}
\end{figure*}

The results for ResNeXt-20 on CIFAR-10 are presented in Fig.~\ref{fig:resnext20_cifar10_sup}.

\subsubsection{ResNet-56}

\begin{figure*}[!h]
    \centering
    \includegraphics[width=0.9\textwidth]{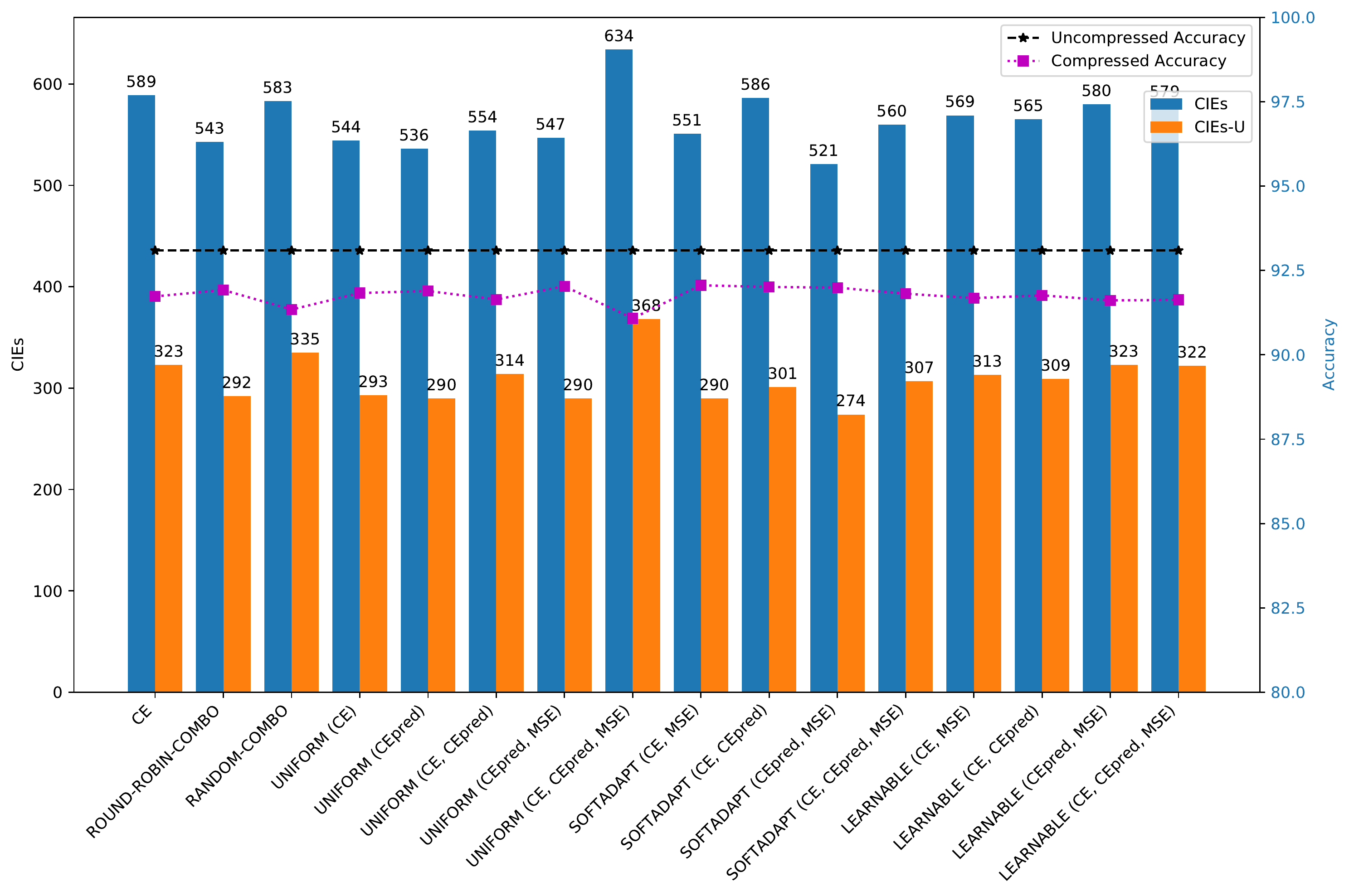}
    \caption{Performance of each loss+optimizer combination on ResNet-56 (CIFAR-10)}
    \label{fig:resnet56_cifar10_sup}
\end{figure*}

The results for ResNet-56 on CIFAR-10 are presented in Fig.~\ref{fig:resnet56_cifar10_sup}.

\clearpage

\subsubsection{ResNet-164}

\begin{figure*}[!h]
    \centering
    \includegraphics[width=0.9\textwidth]{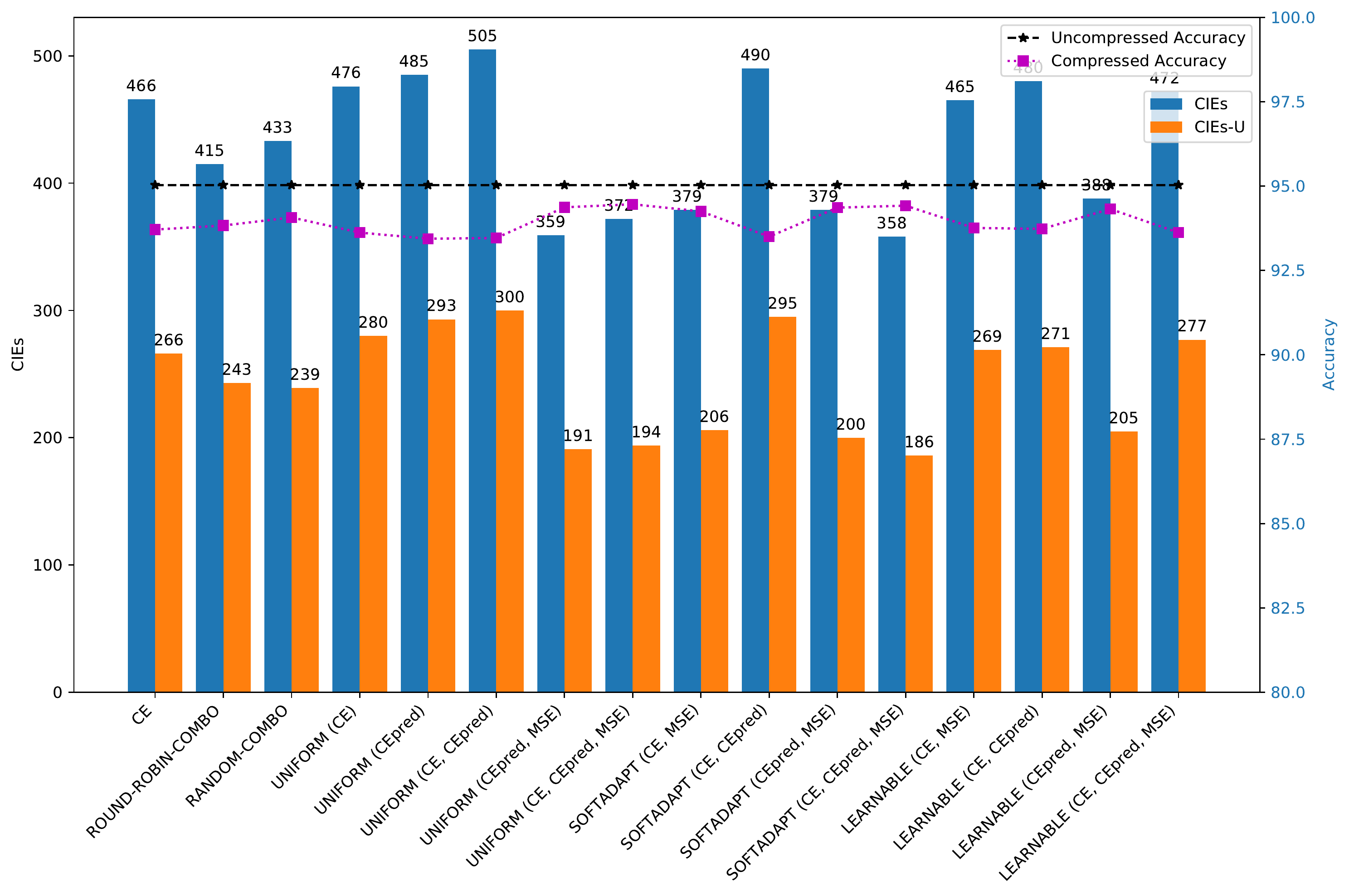}
    \caption{Performance of each loss+optimizer combination on ResNet-164 (CIFAR-10)}
    \label{fig:resnet164_cifar10_sup}
\end{figure*}

The results for ResNet-164 on CIFAR-10 are presented in Fig.~\ref{fig:resnet164_cifar10_sup}.

\subsubsection{ResNeXt-164}

\begin{figure*}[!h]
    \centering
    \includegraphics[width=0.9\textwidth]{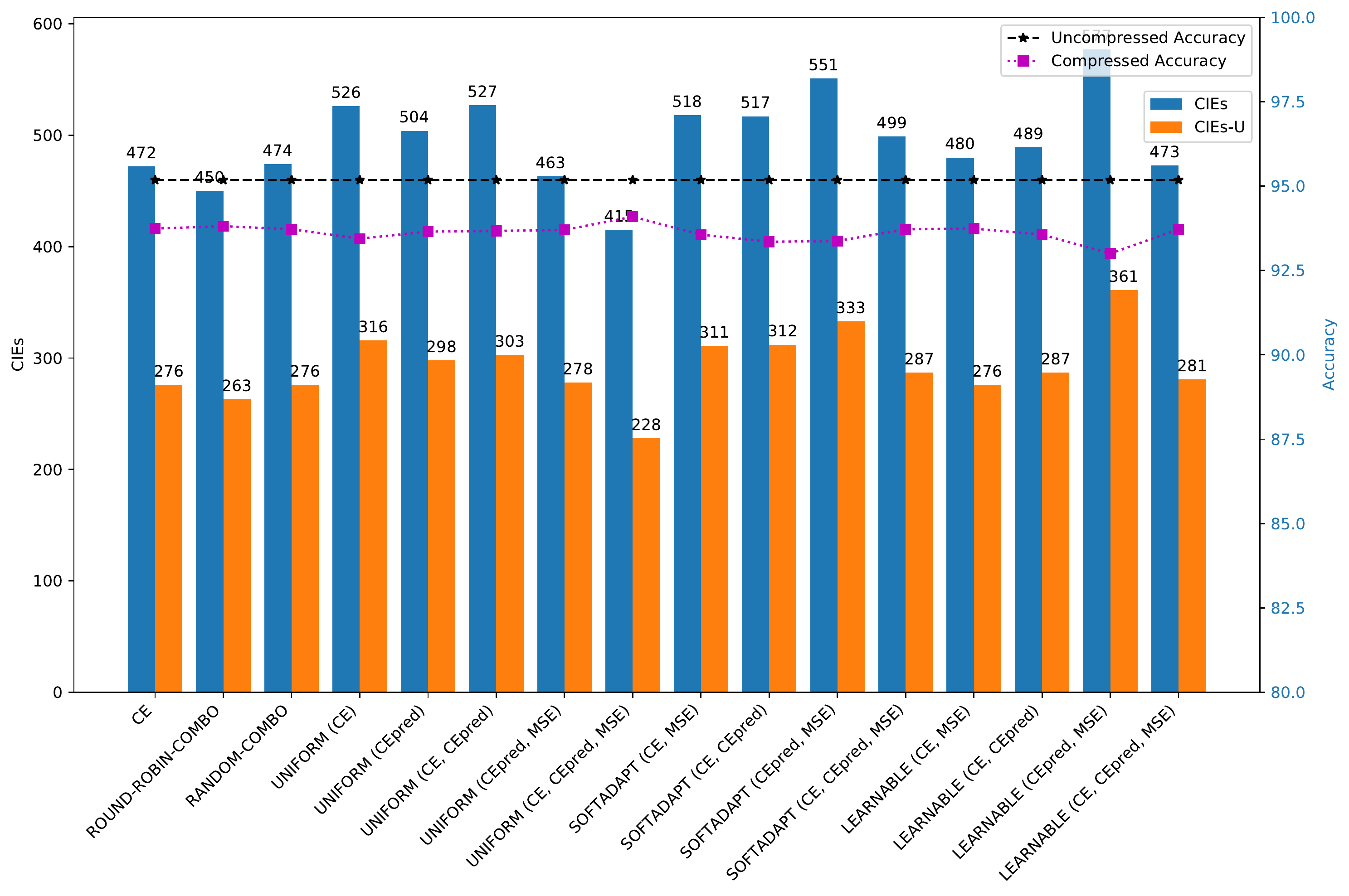}
    \caption{Performance of each loss+optimizer combination on ResNeXt-164 (CIFAR-10)}
    \label{fig:resnext164_cifar10_sup}
\end{figure*}

The results for ResNeXt-164 on CIFAR-10 are presented in Fig.~\ref{fig:resnext164_cifar10_sup}.

\clearpage

\subsection{Results on CIFAR-100}

\subsubsection{ResNet-20}

\begin{figure*}[!h]
    \centering
    \includegraphics[width=0.9\textwidth]{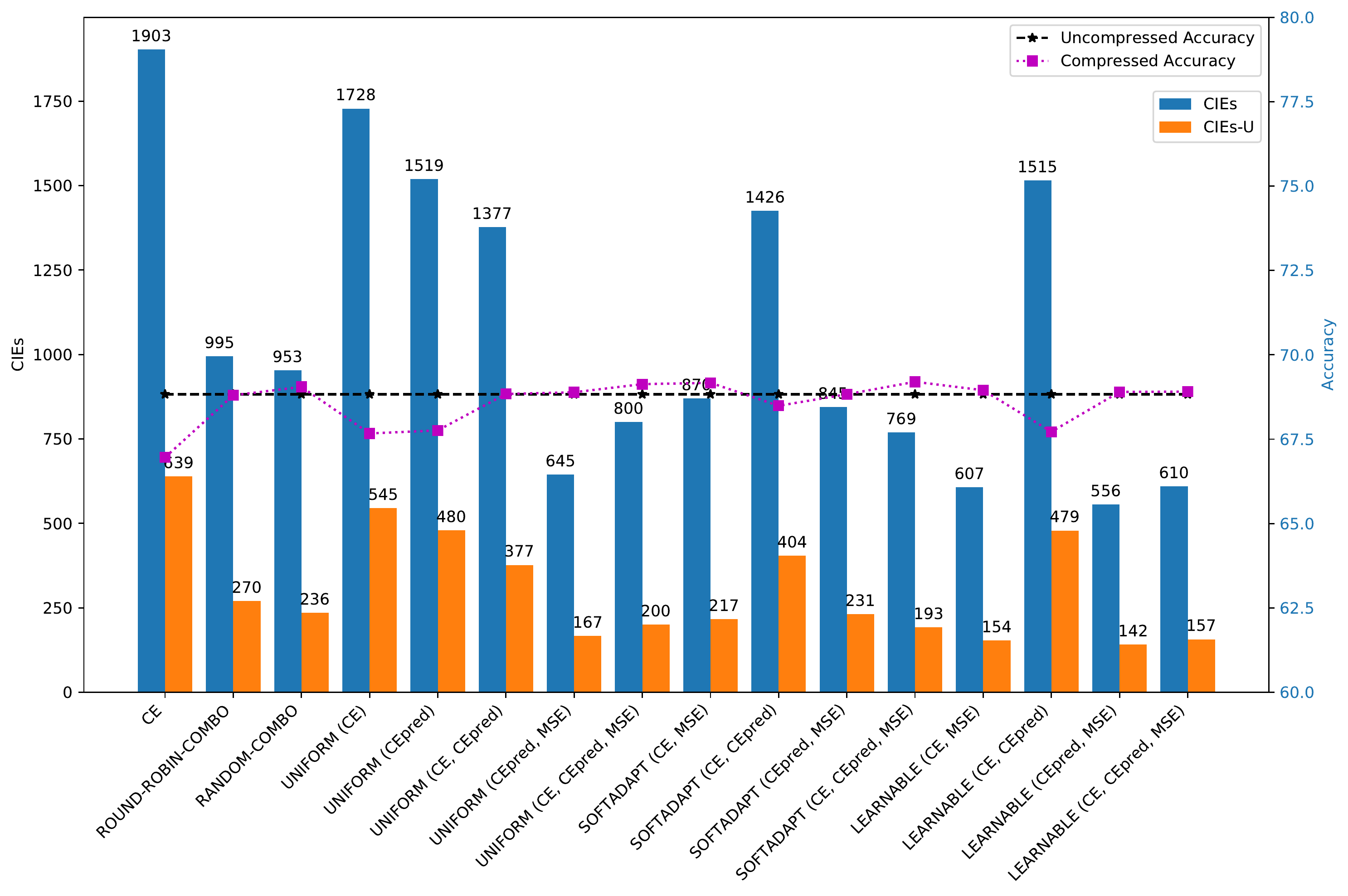}
    \caption{Performance of each loss+optimizer combination on ResNet-20 (CIFAR-100)}
    \label{fig:resnet20_cifar100_sup}
    \vspace*{-6mm}
\end{figure*}

The results for ResNet-20 on CIFAR-100 are presented in Fig.~\ref{fig:resnet20_cifar100_sup}.

\subsubsection{ResNeXt-20}

\begin{figure*}[!h]
    \centering
    \includegraphics[width=0.9\textwidth]{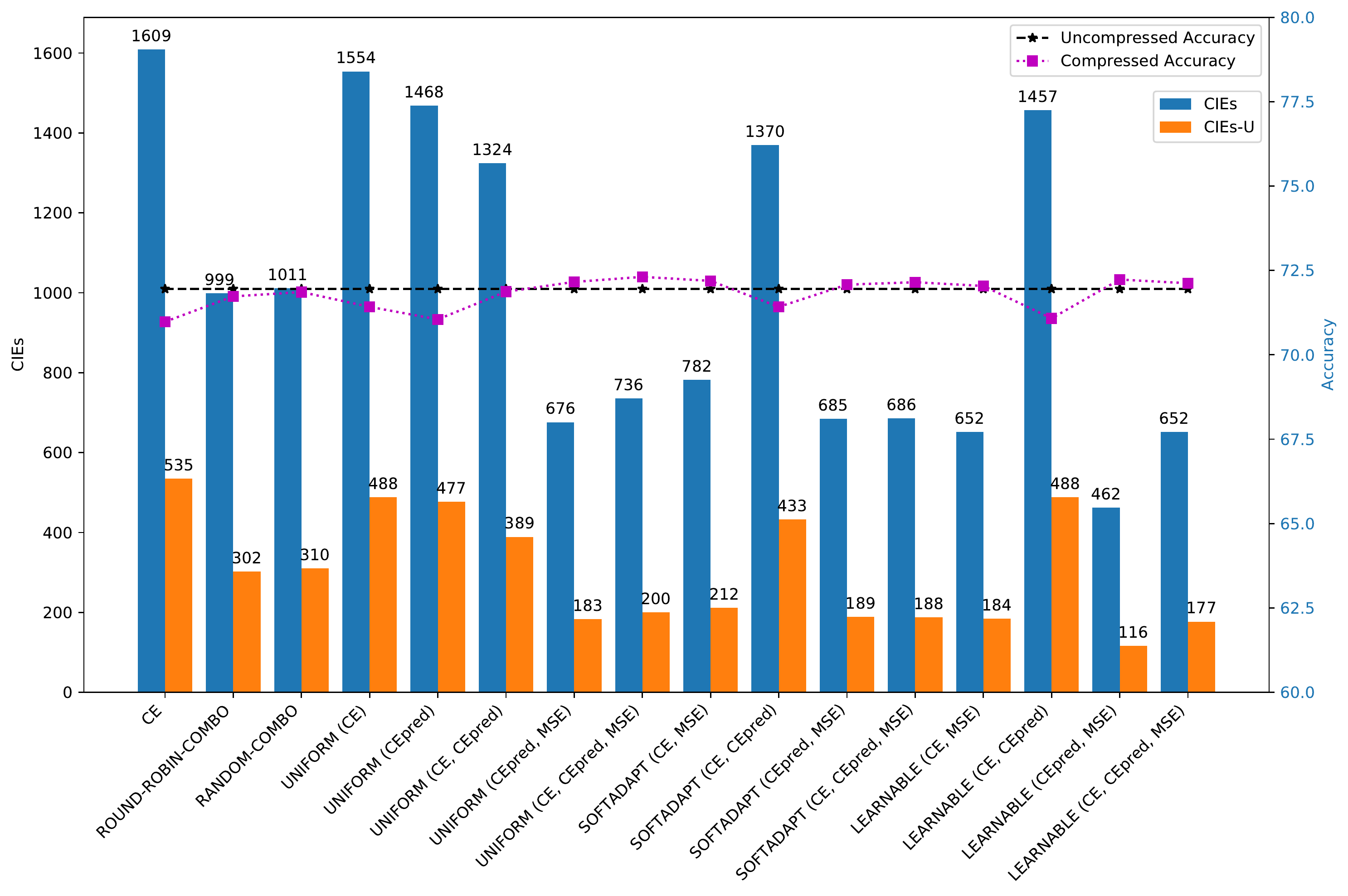}
    \caption{Performance of each loss+optimizer combination on ResNeXt-20 (CIFAR-100)}
    \label{fig:resnext20_cifar100_sup}
\end{figure*}

The results for ResNeXt-20 on CIFAR-100 are presented in Fig.~\ref{fig:resnext20_cifar100_sup}.

\clearpage

\subsubsection{ResNet-164}

\begin{figure*}[!h]
    \centering
    \includegraphics[width=0.9\textwidth]{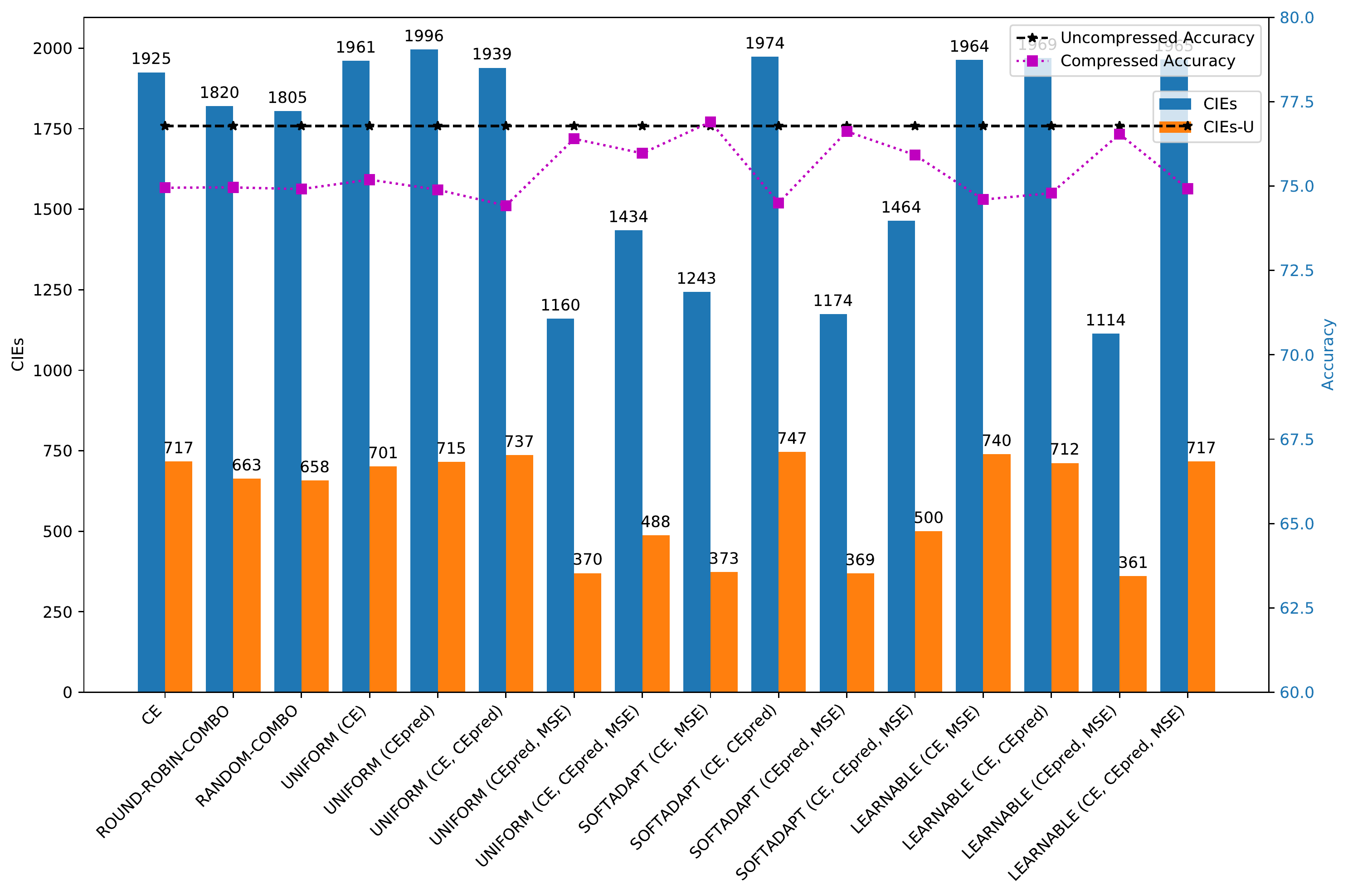}
    \caption{Performance of each loss+optimizer combination on ResNet-164 (CIFAR-100)}
    \label{fig:resnet164_cifar100_sup}
\end{figure*}

The results for ResNet-164 on CIFAR-100 are presented in Fig.~\ref{fig:resnet164_cifar100_sup}.

\subsubsection{ResNeXt-164}

\begin{figure*}[h]
    \centering
    \includegraphics[width=0.9\textwidth]{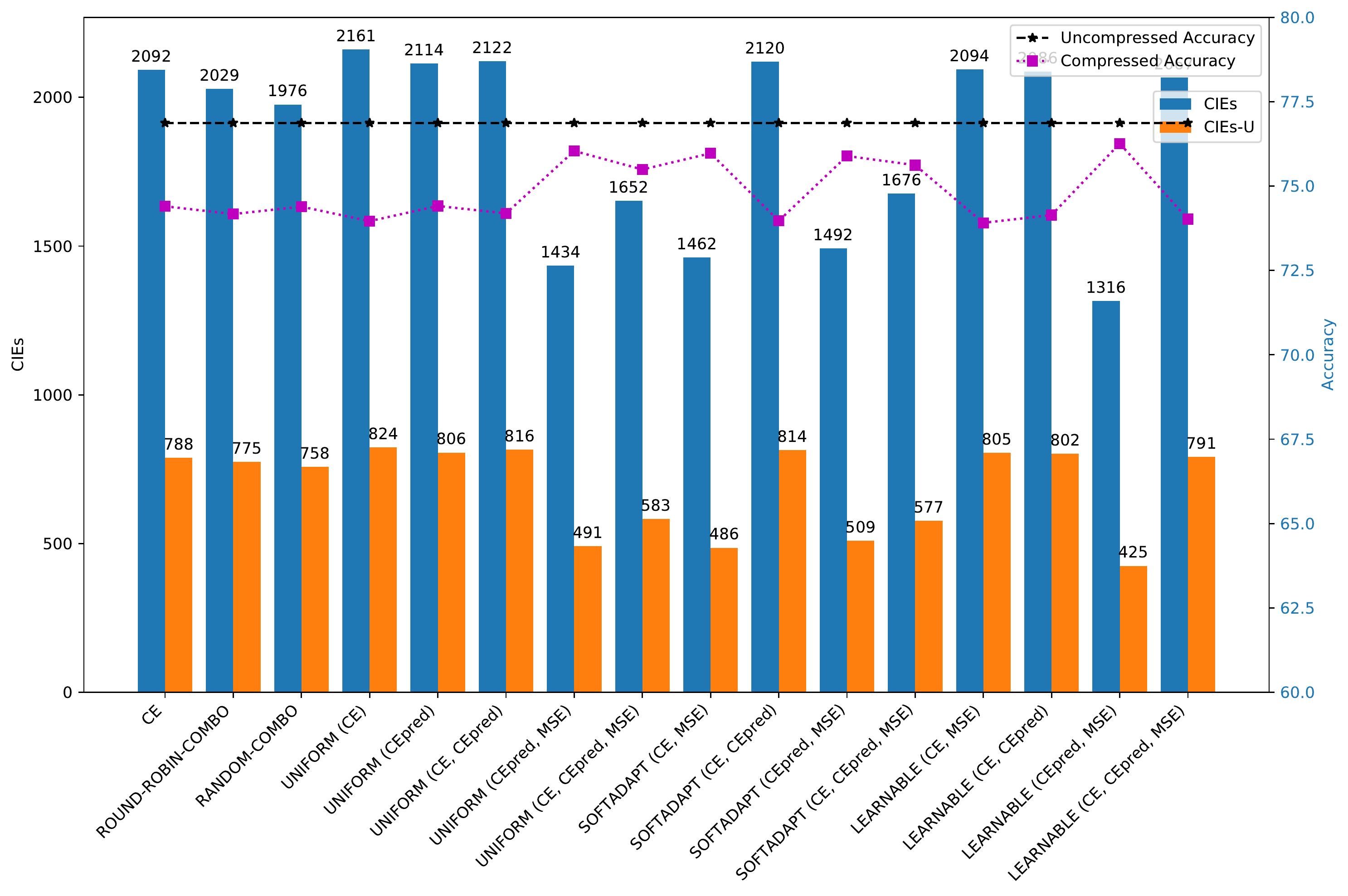}
    \caption{Performance of each loss+optimizer combination on ResNeXt-164 (CIFAR-100)}
    \label{fig:resnext164_cifar100_sup}
\end{figure*}

The results for ResNeXt-164 on CIFAR-100 are presented in Fig.~\ref{fig:resnext164_cifar100_sup}.

\clearpage

\end{document}